\documentclass[times,twocolumn,final]{elsarticle}

\usepackage{amssymb}
\usepackage{latexsym}

\usepackage{url}
\usepackage{xcolor}
\usepackage{caption}
\usepackage{hyperref}
\usepackage{etoolbox}
\definecolor{newcolor}{rgb}{.8,.349,.1}
\journal{Medical Image Analysis}

\usepackage{natbib}
\usepackage{graphicx}
\usepackage{soul}
\usepackage{mathrsfs}
\usepackage{calrsfs}
\usepackage{amsmath}
\usepackage{amssymb}
\usepackage{pifont}
\newcommand{\xmark}{\ding{55}}%
\DeclareMathAlphabet{\pazocal}{OMS}{zplm}{m}{n}
\newcommand{\Lb}{\pazocal{L}}
\sethlcolor{green}
\newcommand{\minus}{\scalebox{0.75}[1.0]{$-$}}
\newcommand{\plus}{\scalebox{0.75}[1.0]{$+$}}

\begin{document}


\begin{frontmatter}

\title{An Uncertainty-aware Hierarchical Probabilistic Network for Early Prediction, Quantification and Segmentation of Pulmonary Tumour Growth}


\author[1,2]{Xavier Rafael-Palou\corref{cor1}}
\cortext[cor1]{Corresponding author}
\ead{xavier.rafael@eurecat.org}

\author[3]{Anton Aubanell}

\author[2]{Mario Ceresa}

\author[1]{Vicent Ribas}

\author[2]{Gemma Piella}

\author[2,4]{Miguel A. González Ballester}

\address[1]{Eurecat Centre Tecnològic de Catalunya, eHealth Unit, Barcelona, Spain}
\address[2]{BCN MedTech, Dept. of Information and Communication Technologies, Universitat Pompeu Fabra, Barcelona, Spain}
\address[3]{Vall d'Hebron University Hospital, Barcelona, Spain}
\address[4]{ICREA, Barcelona, Spain}

\begin{abstract}

Early detection and quantification of tumour growth would help clinicians to prescribe more accurate treatments and provide better surgical planning. However, the multifactorial and heterogeneous nature of lung tumour progression hampers identification of growth patterns. In this study, we present a novel method based on a deep hierarchical generative and probabilistic framework that, according to radiological guidelines, predicts tumour growth, quantifies its size and provides a semantic appearance of the future nodule. Unlike previous deterministic solutions, the generative characteristic of our approach also allows us to estimate the uncertainty in the predictions, especially important for complex and doubtful cases. Results of evaluating this method on an independent test set reported a tumour growth balanced accuracy of 74\%, a tumour growth size MAE of 1.77 mm and a tumour segmentation Dice score of 78\%. These surpassed the performances of equivalent deterministic and alternative generative solutions (i.e. probabilistic U-Net, Bayesian test dropout and Pix2Pix GAN) confirming the suitability of our approach. 


\end{abstract}

\begin{keyword}
\MSC 41A05\sep 41A10\sep 65D05\sep 65D17

Lung cancer\sep Tumour growth\sep Uncertainty\sep Deep learning
\end{keyword}

\end{frontmatter}

\section{Introduction}



Pulmonary nodule malignancy is usually assessed based on relatively few parameters such as longest axial diameter, tumour growth and time between observations\footnote{https://my.clevelandclinic.org/health/diseases/14799-pulmonary-nodules}. Depending on these values and the recommendations made by international radiological guidelines \citep{macmahon2017guidelines}, experts make conjectures and draw their conclusions. From the different malignancy parameters, pulmonary tumour growth is one of the most important indicators when assessing lung cancer by computed tomography (CT) \citep{van2009management}. In particular, clinicians commonly assess tumour growth by imaging surveillance, measuring the nodule diameter along different CT studies taken at different time-points \citep{bankier2017recommendations}. 

Anticipating the tumour growth rate would help clinicians to prescribe more accurate tumour treatments and surgical planning. 
However, lung tumours are highly heterogeneous (e.g. in size, texture and morphology) and their assessment is subject to inter and intra-observer variability (up to 3 mm in diameter on spiculated nodules \citep{han2018influence}), making it complex to derive general patterns of tumour growth. 

Due to the importance of supporting clinicians in this task, several efforts have been done from the computer vision and artificial intelligence community. Traditionally, the tumour growth prediction problem has been addressed through complex and sophisticated mathematical models \citep{sarapata2014comparison}, such as those based on the reaction-diffusion equation \citep{talkington2015estimating, swanson2002quantifying} also known as Fisher–Kolmogorov model. These methods provide informative results and explainability. However, the number of model parameters is often limited (e.g. 5 in \cite{wong2015tumor}), which might not be sufficient to model the inherent complexities of the growing patterns of the tumours. 

Recently, deep learning and in particular deep convolutional neural networks (CNN) have shown a great ability to automatically extract high-level representations from image data \citep{najafabadi2015deep}. 
This has enabled performance improvements over conventional approaches in various medical imaging problems, such as nodule detection \citep{setio2017validation}, segmentation \citep{messay2015segmentation}, re-identification \citep{rafael2020re} and malignancy classification \citep{ciompi2017towards}.  

 Tumour growth estimation has also been addressed with deep learning for brain, pancreatic and/or colorectal cancer, using data from longitudinal CT/PET or magnetic resonance imaging (MRI) \citep{zhang2017convolutional,zhang2019spatio,katzmann2018predicting}. Proposed deep architectures usually rely on CNNs and recurrent neural networks (RNN) \citep{hochreiter1997long}, for extracting spatial and temporal tumour growth patterns and correlations. Recently, generative networks such as those based on adversarial learning \citep{goodfellow2014generative} and variational auto-encoders \citep{kingma2013auto} have also been proposed to enhance grow prediction and clinical interpretability by estimating future images of the tumour \citep{elazab2020gp,petersen2019deep}. 
 
Few works have tackled lung tumour growth estimation \citep{li2020learning, wang2019toward}. In \cite{li2020learning}, they proposed two 3D CNNs to extract warping and texture patterns to predict malignancy risk and future aspects of the tumour.  
In contrast, in \cite{wang2019toward}, they proposed a CNN combined with a RNN extended with an attention mechanism \citep{luong2015effective} to find temporal patterns to provide trajectories of lung tumour evolution using MRI images. We provide further details of these recent works in section 2. These works, however, address the problem of growth prediction in a deterministic way, providing a single prediction without considering uncertainties. Therefore, the models do not usually take into account neither the variability in the annotations of the experts, nor the risk of failure. This could partially explain why in clinical settings the credibility of these models is questioned and their adoption limited.  

Along with the recent interest on tumour growth prediction and uncertainty with deep learning, this work aims to take a step forward in these promising research directions. In particular, we propose a probabilistic-generative model able to predict, given a single time-point image of the lung nodule, multiple consistent structured output representations. To do this, the network learns to model the multimodal posterior distribution of future lung tumour segmentations by using variational inference and injecting the posterior latent features. Eventually, by applying Monte-Carlo sampling on the outputs of the trained network, we estimate the expected tumour growth mean and the uncertainty associated with the prediction. 


The contribution of this work is three-fold. First, to the best of our knowledge, this is the first time pulmonary nodule growth is estimated using deep learning and nodule diameter annotations from multiple experts. Second, this is the first time that model uncertainty is reported using a deep learning approach to predict lung nodule growth. Third, a new deep learning solution is presented, building on an existing hierarchical generative and probabilistic segmentation framework, for lung nodule growth prediction, quantification and visualization. 



The rest of the article is organized as follows. Section 2 describes the most recent works on tumour growth estimation. Section 3 details the proposed method for modeling lung tumour growth and its related uncertainty. Sections 4 and 5 report and discuss the experimental results of applying our approach and other competing solutions, on a longitudinal cohort. Finally the conclusions are summarized in Section 6.

\section{Related work}

\subsection{Deep learning deterministic approaches}

Deep learning, and in particular CNNs, seems a perfect match for leveraging tumour growth for its intrinsic capability of automatically extracting deep representations and correlations between multiple images \citep{bengio2017deep}. 


One of the earliest deep learning studies addressing tumour growth estimation was for pancreatic cancer \citep{zhang2017convolutional}. The authors proposed the use of two (invasion/expansion) stream CNNs, relying on 2D patch images of the tumour, for predicting future tumour segmentations as well as tumour volume growth rates. Interestingly, the method allowed integration with clinical data to enable personalization. Best method performances achieved 86\% of Dice score and 8.1\% relative volume difference (RVD). Those overcame state-of-the-art of conventional mathematical models \citep{wong2016pancreatic} for that disease type. However, the size of the test set was too small (10 cases) to extract robust conclusions. Also, to make inference this network required multimodal images (i.e. dual phase contrast-enhanced CT and FDG-PET), as well as three time points spanning between three to four years, which represented strong pre-conditions for the usability of the model.

Aiming to go beyond black-box predictions for lung tumour malignancy \citep{ciompi2017towards,huang2019prediction}, recent work \citep{li2020learning} proposed a method to generate a future image of the nodule. To do this, a temporal module encoded the distance at which to make the prediction, and two 3D U-Net \citep{milletari2016v,ronneberger2015u} networks extracted the warped and texture image features of the lung nodule. The network was trained with more than 300 pairs (prior and current studies) of 3D nodule centered patches. Experiments reported a high balanced accuracy score of 86\% for nodule progression, although a relative Dice score of 65\% for future nodule segmentation.  
The gap in the model's ability to provide future segmentations of the tumours, the use of a tailored criteria to determine nodule growth instead of conventional metrics (e.g. longest diameter or double time volume) or not taking into account inter-observer variability, shows the need to continue with the investigation of more reliable and effective solutions.  

An alternative approach, especially suitable for temporal series, are the RNNs, in particular the Long Short-Term Memory (LSTM) networks \citep{hochreiter1997long}. They were designed for the next time-step status prediction in a temporal sequence capable of learning long-term dependencies. Some recent works have used this type of architectures for tumour growth prediction. For instance, in \cite{zhang2019spatio}, a 3D convolutional LSTM network \citep{shi2015convolutional} was proposed for predicting pancreatic tumour growth. Interestingly, in this study, features from the clinical history of the patient were integrated in the network with the intention to find extra non-linear relationships between spatial and temporal features. This approach used a limited dataset (33 cases) and required having series ($\geq$ 2) of previous images of the lesion, which for early tumour growth estimation is not the best scenario due to the aggressiveness of the disease. Regarding lung tumour growth, in \cite{wang2019toward} a network was proposed to combine convolutional layers and gated recurrent units with an attention mechanism \citep{luong2015effective}. The goal was to predict spatial and temporal trajectories over a course of radiotherapy using a longitudinal MRI dataset. Although the purpose of this study is similar to ours (i.e. future lung tumour growth estimation), the complexity of the problem differs in that the images analyzed were MRI (instead of CT), the period of the predictions were weeks (instead of months/years), and the number of input images (i.e. 2-3) to the network was larger than in our case. 


\subsection{Deep generative networks}

Another way to tackle tumour growth prediction is by using deep generative models. One of the most popular is generative adversarial networks (GAN) \citep{goodfellow2014generative}. This framework consists of two networks, the generator and the discriminator, that compete with each other in a zero-sum game where the generator aims to increase the error rate of the discriminator network. Thus, the generator learns to map points from a latent space, usually sampled from a multivariate standard normal distribution, into observations that look as if they were sampled from the original dataset. The discriminator tries to predict whether an observation comes from the original dataset.

GANs have been recently applied to predict future tumour/disease growth over time. For instance, in  \cite{li2020dc} they proposed a 2D deep convolutional GAN for discriminating between true tumour progression and pseudo-progression of glioblastoma multiforme. The results confirmed its suitability for prediction and feature extraction, although only one image per tumor was used in the study. In \cite{elazab2020gp} they built a stacked 3D GAN for growth prediction of gliomas using temporal evolution of the tumour. 
Although high performances were reported (88\% Dice score), the database was composed by only 18 subjects, in which all tumours always grew. In \cite{rachmadi2020automatic}, they compared different GAN networks to predict the evolution of white matter hyper\-intensities. They also demonstrated the potential of using GANs in a semi-supervised scheme, improving results of a deterministic U-ResNet \citep{zhang2018road}. Despite the satisfactory performances obtained with GANs, this type of network suffers from mode collapse\citep{goodfellow2016nips}, that is, they hardly generate correct representations of the probability output distribution, so they may not be adequate to model uncertainty.


Another well-known approach for addressing image generation is deep auto-encoders (AE). This framework uses an encoder which embeds the input into a representation vector, and a decoder, which projects the vector back to the original manifold.  The representation vector is a compression of the original image into a lower dimensional, latent space. The idea is that, by choosing any point in a latent space, a novel image is generated by passing this point through a decoder (as it learned to convert points, or representations, in a latent space into viable images). Therefore, the learning process of this network consists on minimizing the reconstruction error, which is the error between the original image and the reconstruction from its representation. Since auto-encoders do not force continuity in space, images are poorly generated at sampling time. 

One successful extension from auto-encoders are variational auto-encoders (VAE) \citep{kingma2013auto,rezende2014stochastic}. In particular, the encoder retrieves two vectors, the mean and log-variance vectors, which together define a multivariate distribution in the latent space.  When a random point is sampled from this distribution, the decoder produces a similar image, guaranteeing the continuity in the latent space. The way to achieve this, 
is by making the output distribution of the encoder as close as possible to a standard multivariate normal distribution using the Kullback-Leibler divergence (KL) loss. Thus, the total loss function of the VAE is composed by the sum of the KL-divergence loss and the reconstruction loss. A variant of VAEs was created to generate multiple outputs from a single input. Precisely, conditional variational auto-encoders (CVAE) \citep{sohn2015learning} were proposed to model the distribution of a high dimensional space as a generative model conditioned on the input. Therefore, the prior on the latent variable is conditioned by the input.

Few works have applied auto-encoders and their variants for tumour/disease growth prediction. In \cite{katzmann2018predicting} they proposed using a deep auto-encoder attached to a fully connected network architecture for colorectal tumour growth detection. Despite providing results close to the RECIST methodology\footnote{https://recist.eortc.org/} and radiomic measures, the use of the auto-encoder was for mere feature reduction. In \cite{basu2019early}, the authors applied a VAE for progression of Alzheimer disease from structural MRI images. Their experiments demonstrated that VAE outperforms conventional CNNs on doubtful cases as it acts as a soft classifier learning a Gaussian distribution. Also, for patient risk analysis they observed that VAE produced less false positive cases, sampling from the latent space, than determininstic CNNs. However, CNNs provided better overall performances. In another study \citep{ravi2019degenerative}, they conditioned a deep auto-encoder on fixed characteristics like age and diagnosis, to generate sequences of 3D MRI for Alzheimer's disease progression. Despite results outperformed previous 2D versions, some artefacts and false structures were noted on the generated images. Moreover, additional terms were required to ensure loss stability, latent space continuity, reducing memory constraints and restoring 3D outputs.  

\subsection{Uncertainty in deep learning}

Contradictorily, given the multifactorial and complex nature of the problem, uncertainty in the prediction of tumour growth was not addressed in any of the aforementioned studies. However, 
uncertainty information about the output of a network could make them safer and more reliable since it would allow indicating potential mis-segmented or low confident regions, or guiding user interactions for refinement of the results. Two common approaches have been proposed for modeling uncertainty in deep learning, Monte Carlo dropout networks (MCDNs) \citep{gal2016bayesian} and Bayesian neural networks (BNNs) \citep{shridhar2019uncertainty}. MCDNs use dropout layers as a Bayesian inference approximation in deep Gaussian processes, and although their implementation is easy, criticism has emerged recently regarding the type of uncertainty that is captured \citep{Osband2016Risk}. BNNs use variational inference to learn the posterior distribution of the weights given a dataset. These weights are implicitly described as (multivariate) probability distributions. This has several consequences. First, it makes the neural network non-deterministic; for every forward pass, we must sample from each weight distribution to obtain a point estimate. Repeated applications of this sampling technique, through Monte Carlo sampling, will result in different predictions which can then be analyzed for uncertainty. Second, it changes the backpropagation algorithm since we cannot flow back the gradients through a sampling operation.

Uncertainty estimation in deep neural networks has been widely investigated for medical image tasks. For instance, in segmentation of multiple sclerosis lesions, some works \citep{nair2020exploring,roy2018inherent} showed that by filtering out predictions with high uncertainty, the models improved lesion detection accuracy. For brain tumour segmentation, other work \citep{eatonrosen2018safe} demonstrated that MCDNs can be calibrated to provide meaningful error bars over estimates of tumour volumes. Moreover, the uncertainty metric based on MCDNs also showed promising results in disease grading of retinal fundal images \citep{leibig2017leveraging, ayhan2018test}. In \cite{lipkova2019personalized} a Bayesian method predicted patient-specific tumour cell densities with credible intervals from high resolution MRI and PET imaging modalities.

Unfortunately, few works have modeled uncertainty for tumour growth estimation. In \cite{petersen2019deep} a deep probabilistic generative model (sPUNet) \citep{kohl2018probabilistic,baumgartner2019phiseg} was used to model glioma growth for radiotherapy treatment planning. The model, based on a combination of a U-Net \citep{ronneberger2015u} and a CVAE \citep{sohn2015learning}, was able to generate multiple future tumour segmentation modes on a given input. Although they demonstrated the potential of providing multiple views over a single solution, they did not report nodule growth performances.

\section{Method}


We present a novel approach to estimate the future growth of pulmonary nodules along with its uncertainty. 
Our approach exploits the generative and probabilistic nature of a recent framework, the hierarchical probabilistic U-Net \citep{kohl2019hierarchical} (HPU), to estimate the output probability distribution of lung nodule growth, conditioned on an initial image of the nodule. Before delving into the details, in the following sub-section we describe the basics of the underlying framework. 


\subsection{Hierarchical probabilistic U-Net}

\begin{figure}[t]
\centering
\includegraphics[scale=.4]{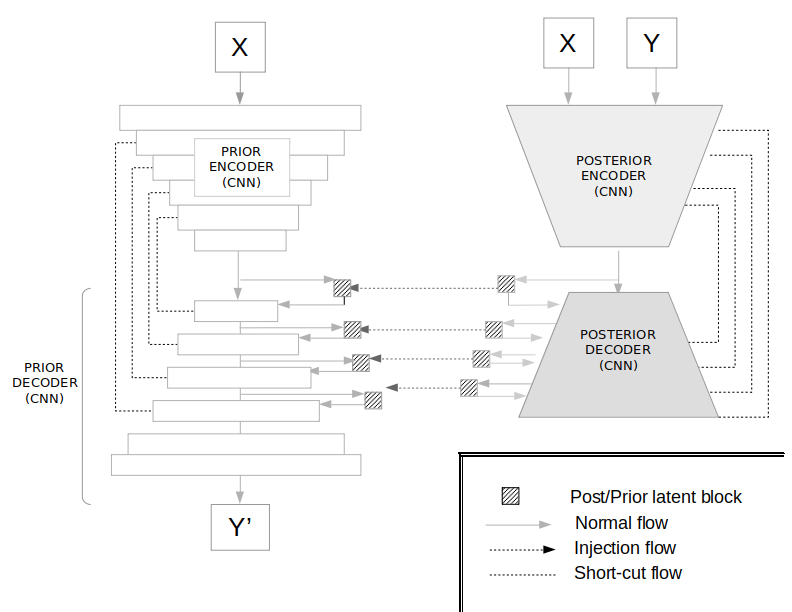}
\caption{General overview of the HPU network architecture. On the left of the picture we can observe the prior network and on the right the posterior. Both networks have different probabilistic latent blocks interleaved along the decoder component.}
\label{fig:hpu_arch}%
\end{figure}

A segmentation framework that provides multiple segmentation instances for ambiguous images was proposed in \cite{kohl2019hierarchical}. This network, schematized in Figure-\ref{fig:hpu_arch}, is composed of two inter-related sub-networks, the posterior and the prior. Both follow a CVAE scheme with a couple of changes. First, the encoder-decoder structure is implemented by a 2D U-Net \citep{ronneberger2015u} extended with residual blocks \citep{he2016identity, drozdzal2016importance} (U-ResNet) and filters adjusted to the input size. Second, instead of a single probabilistic latent block (see Figure-\ref{fig:hpu_updated_dec}) 
at the end of the encoder, several probabilistic latent blocks are interleaved at different levels of the hierarchy of the decoder, to provide fine-grained segmentation samples closer to the ground truth probabilistic distribution. 


The inference process of this network consists on forward-passing an input image, X, through the prior network. Specifically, along the decoder part of the network, feature activation maps are concatenated with vectors, z\textsubscript{i} ($i\leq L$, being L the number of latent hierarchies), obtained from sampling different latent distributions interleaved in the decoder. As a result we obtain a predicted segmentation, Y'. 


The training process of this network aims to pull to each other the prior distribution $p$, encoded by the prior network, and the posterior distribution $q$, defined by the posterior network,  while minimizing the loss of the reconstructed images. This is the same as maximizing the evidence lower bound (ELBO) in variational inference. Therefore, the KL divergence loss ($D\textsubscript{KL}$) between the posterior and the prior distributions is added to the reconstruction objective ($\Lb \textsubscript{rec}$) obtained through the log likelihood (represented by the pixel-wise categorical distribution $P\textsubscript{c}$) between the reconstructed image Y', and the ground truth segmentation Y. Additionally a weighting factor $\beta$, is multiplied to the D\textsubscript{KL} term to balance the overall loss function:
\begin{center}
$
\Lb \textsubscript{ELBO} = \mathop{\mathbb{E}}\textsubscript{z$\sim$Q} [ \minus log P\textsubscript{c}(Y|Y') ] \plus
\beta \sum\limits_{i=0}^{L} \mathop{\mathbb{E}}\textsubscript{z\textsubscript{$<$i}$\sim$Q} D\textsubscript{KL}(q\textsubscript{i}(z\textsubscript{i}|z\textsubscript{$<$i},X,Y)||
p\textsubscript{i}(z\textsubscript{i}|z\textsubscript{$<$i},X))
$
\end{center}
where $\mathop{\mathbb{E}}\textsubscript{z$\sim$Q}$ is the expectation operator, and z a vector sampled from the posterior distribution Q.

\subsection{U-HPNet}

\begin{figure}[t]
\centering
\includegraphics[scale=.35]{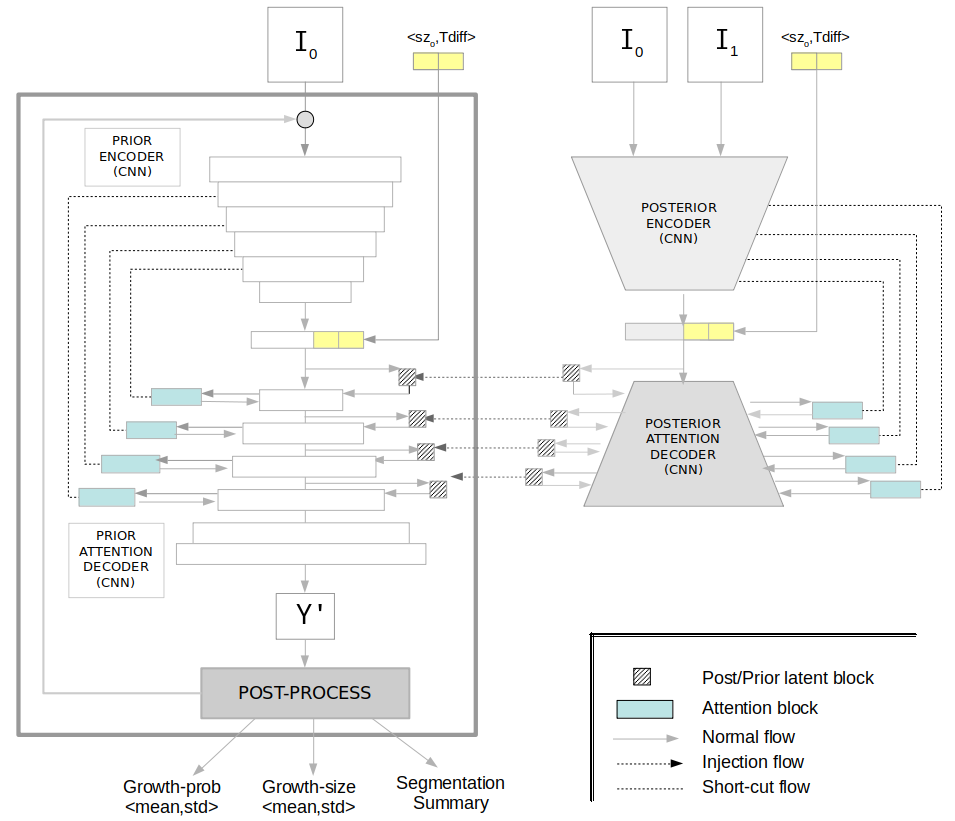}
\caption{General overview of the proposed U-HPNet network architecture. This network is also composed by a prior network (on the left, with further details) and a posterior (on the right). Attatched at the end of the posterior we observe the post-process module aimed at reporting the estimated future growth prediction, size and appearance with the associated uncertainty.}
\label{fig:hpu_updated_arch}%
\end{figure}

Based on the HPU framework, we propose a network (U-HPNet) able to generate plausible future nodule segmentations conditioned on the nodule image, its diameter at time T\textsubscript{0}, and the temporal distance at which to make the prediction. To do this, the U-HPNet uses variational inference to approximate the estimated output distribution to the ground truth, in our case, provided by different graders. Figure-\ref{fig:hpu_updated_arch} shows the overall architecture of the proposed network.

\subsubsection{Architecture}

 Both sub-networks of the U-HPNet (prior and posterior) receive as input an axial nodule image I\textsubscript{0} at time T\textsubscript{0}, while the posterior receives also the axial nodule image I\textsubscript{1} at T\textsubscript{1}. The images are centered patches of 32x32 pixels rather than 128x128 as in the original network. We down-scaled the input size of the network to focus in the relevant parts of the image (i.e. contour and close surrounding of the nodule), and to reduce the number of parameters of the network, especially convenient for small datasets \citep{prasoon2013deep}. 
 Smaller patches were discarded due to the size of the nodules, and larger patches (e.g. 64x64) experimentally did not report any performance gain. 

We conditioned the latent space, learnt by the network, with a couple of extra features:  the time difference (Tdiff) at which to predict nodule growth, and the diameter size (sz\textsubscript{0}) of the nodule image at time T\textsubscript{0}. Tdiff is an ordinal value representing the main time-elapses defined by radiological guidelines (i.e. 6, 12, 24 or more months) \citep{macmahon2017guidelines}. Sz\textsubscript{0} is a numerical value provided (in our case) by radiologists to better estimate the tumour growth. 
In particular, with this feature we aimed to facilitate the network to learn the intrinsic patterns followed by the experts when measuring tumours from the images. 
Both features (Tdiff, sz\textsubscript{0}) were normalized between 0 and 1, and concatenated with the encoder output. 

\begin{figure}[!ht]
\centering
\includegraphics[scale=.35]{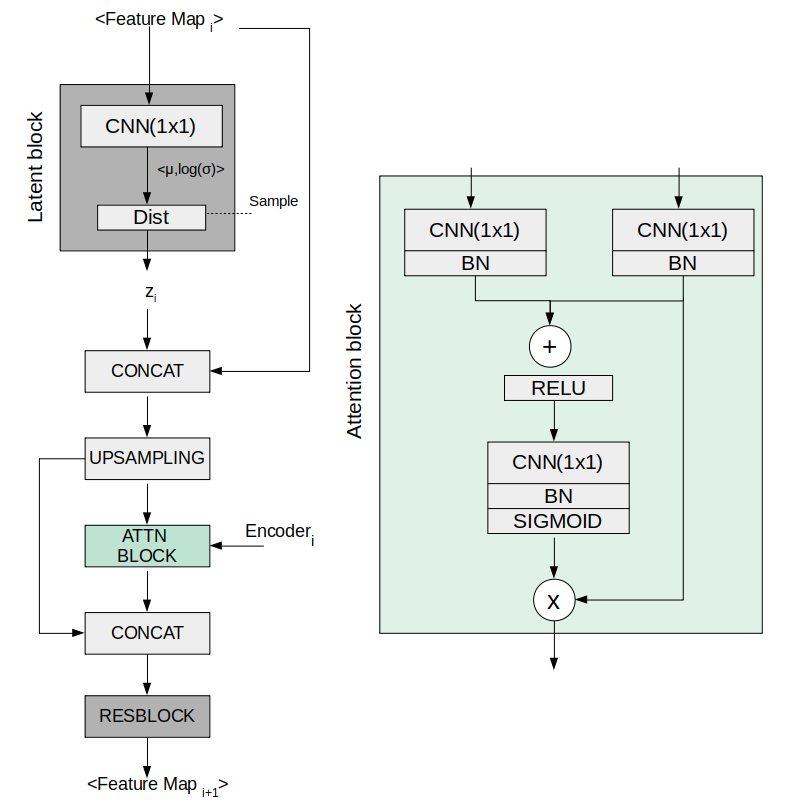}
\caption{On the left we show a more detailed view of the different components of a decoder layer of the U-HPNet. On the right we see the elements that compose an attention block.}
\label{fig:hpu_updated_dec}%
\end{figure}

Regarding the network architecture, both sub-networks use the same 2D U-ResNet as in the original HPU, but adapted to the proposed input size (32x32). Also, up to 4 prior/posterior latent blocks are interleaved in the decoder of the sub-networks, generating latent feature vectors (z) of 1, 4, 16, 64  dimensions respectively.

Additionally, we integrated a soft attention mechanism in the decoder part of the sub-networks with the intention of detecting small and minor changes in the structure of the nodule images. To do this, we followed a recent work \citep{oktay2018attention} in which a grid-attention mechanism was integrated in a U-ResNet. The attention mechanism aims at progressively suppressing feature responses in irrelevant background regions. To do this, attention gates are integrated before the concatenation operation to merge only relevant activations. Figure-\ref{fig:hpu_updated_dec} provides further details regarding the components of the attention mechanism and how it was integrated in the decoder of the sub-networks. 

\subsubsection{Loss function}

On the conventional ELBO loss function used in the original HPU paper, we incorporated a couple of modifications in the reconstruction loss ($\Lb \textsubscript{rec}$) term.
In particular, we used the L1 distance between the predicted $D1'$ and the ground truth $D1$ tumour diameters, and the intersection over union (IoU) between the predicted $Y'$ and ground truth $Y$ tumour segmentation. Also a weighting ($\gamma$) factor was used on the combined loss to balance the ranges of both terms.
\begin{center}
$\Lb \textsubscript{rec} = \Lb \textsubscript{IoU}(Y,Y') + \gamma \Lb \textsubscript{L1} (D1,D1')$
\end{center}
We used the L1 loss to prioritize the diameter fidelity, and consequently improve network performance. Also, we used $IoU$ loss as a good approximation function when learning on imbalanced data conditions \citep{oksuz2020imbalance}, which in our case was caused by having a much smaller number of pixels belonging to the tumour than to the background. In our experiments, we found better performances setting $\gamma$ to one.

\subsubsection{Post-processing}

The generative ability of the proposed network offers the possibility to produce future nodule segmentations, sampling from the latent space and injecting the resulting vectors in the network, for a given input. This may be useful from a medical exploratory point of view, but for practical reasons a more useful outcome should be presented to the clinicians. To this end, we formulated a generic and embeddable post-processing module that converts multiple predicted segmentations into a lung nodule growth prediction, size and segmentation visualization with the uncertainty associated to each of them. 
Precisely, the post-processing module applies Monte-Carlo sampling by running the network $K$ times (K=1000) with the same input image. In particular, for each iteration, a sample from all the hierarchical latent blocks of the prior network is injected in the corresponding location of the decoder part of the (prior) network, to produce a new segmentation. As a result we obtained $K$ random nodule segmentations. For each predicted segmentation, we extracted its longest diameter $D1'$, using conventional image processing libraries. With the vector of $K$ nodule diameters, we computed the vector of predicted nodule growths, $\Delta$, by subtracting the input nodule diameter size $D0$ (the aforementioned sz\textsubscript{0}) to the predicted diameters $D1'$. From the resulting vector $\Delta$ of predicted nodule growths, we computed its mean and standard deviation as measures of nodule growth size and its associated uncertainty.

In addition, we computed the probability that the nodule growth is at least of 2 mm (threshold recommended in clinical guidelines for tumour growth \citep{macmahon2017guidelines}). For this, for each of the K nodule growths, we used the logistic function $f(\Delta\textsubscript{i})=1/(1+e^{-\Delta\textsubscript{i}+2})$.
From the resulting K-length vector of probabilities, we considered the mean and the standard deviation as the estimated nodule growth probability and its associated uncertainty.


Finally, the post-processing module also outputs two images, both corresponding to the predicted future tumour appearance (at T\textsubscript{1}). In particular, and inspired by \citep{kendall2015bayesian}, one of the images is the per-pixel mean of all $K$ predicted segmentations and the other the per-pixel standard deviation.

\subsection{Comparison with related works}

\begin{figure}[!ht]
\centering
\includegraphics[scale=.4]{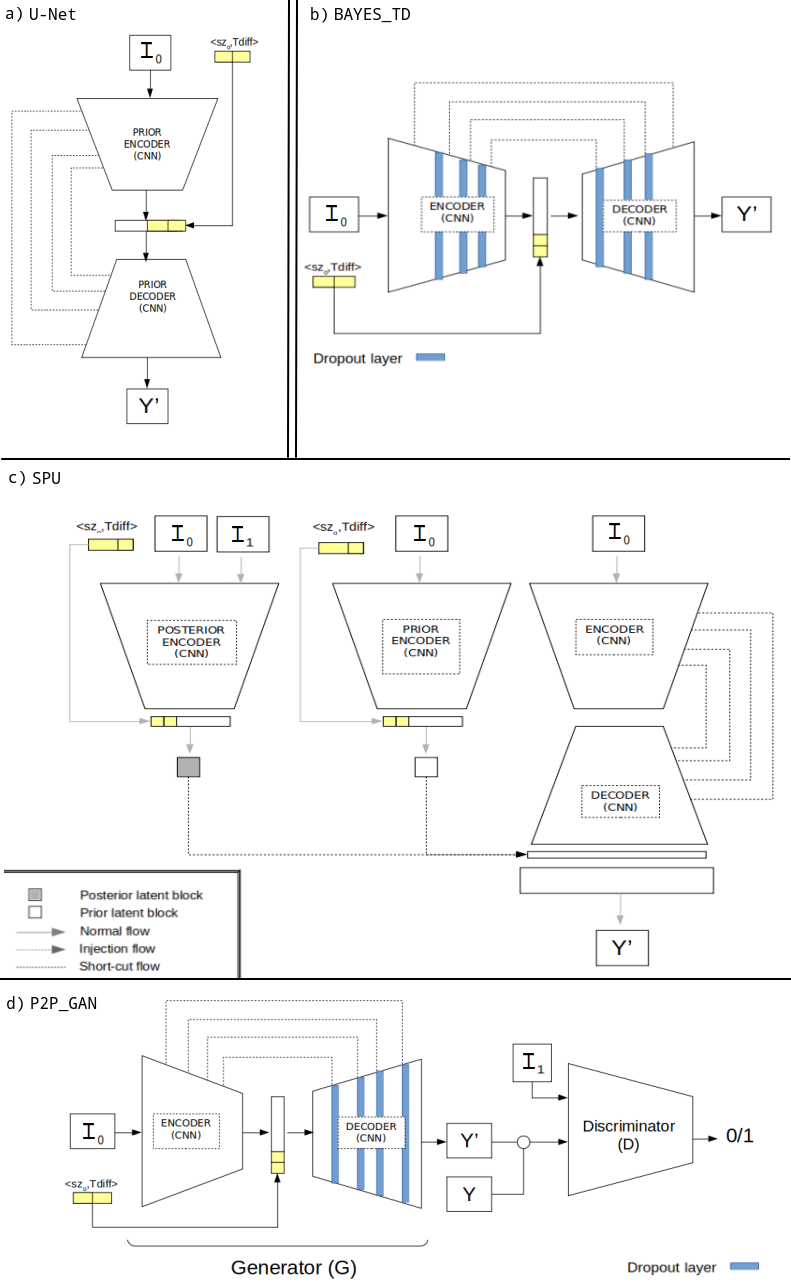}
\caption{Four alternative network architectures proposed for lung nodule growth estimation. At the top we have the U-ResNet and the generative Bayesian dropout. In the center, we show the probabilistic U-Net. Below we find the Pix2Pix cGAN network proposed. }
\label{fig:arch_all}%
\end{figure}

Since we did not find any other deep generative network to provide lung tumour growth predictions and their associated uncertainty, we adapted 4 different state-of-the-art deep architectures to compare the performance of our method (see Figure-\ref{fig:arch_all}), one deterministic network and three generative.  

To allow a fair comparison, all these networks had the same U-ResNet backbone proposed for the U-HPNet, with same number of layers and filters. Also these models were configured with same data augmentation, optimization algorithm, batch size and learning rate than the U-HPNet network. Moreover, these networks had the same input (I\textsubscript{0}, sz\textsubscript{0} and Tdiff) and output as the U-HPNet (i.e. an estimated future segmentation of the nodule). For the non-deterministic models, the output was post-processed to evaluate tumour growth prediction, diameter growth and the segmentation performance.

As for the deterministic (or baseline) approach, we used a single U-ResNet like network, Figure-\ref{fig:arch_all}a. This network was trained using a conventional loss function, formed by a pixel-wise binary cross entropy, without any additional configuration. 

The first generative method consisted on a Bayesian dropout network (BAYES$\_$TD) following the Bayesian SegNet proposed in \cite{kendall2015bayesian}. This approach provides a probabilistic pixel-wise semantic segmentation by enabling dropout at inference time. Therefore this approach aims to find the posterior distribution over the convolutional weights, W, given the observed image I\textsubscript{0} and labels Y, i.e. $p(W|I\textsubscript{0},Y)$. According to the authors, the best configuration was obtained using dropout in the central part of the network. Thus, we followed the same suggestion and we setup dropout (p=0.5) layers in the 3 last encoder and 3 initial decoder blocks of the U-ResNet, Figure-\ref{fig:arch_all}b. This network was trained using pixel-wise binary cross entropy. Hence, we used dropout at inference time as a way to get samples from the posterior distribution. 

The second proposed generative network was the former version of the HPU, the standard probabilistic U-Net (SPU) \citep{kohl2018probabilistic}. This approach goes beyond the notion of reporting a per-pixel probability map, by capturing the co-variances between pixels and providing consistent structured outputs. To do this, two networks: the prior (having as input a nodule I\textsubscript{0}) and the posterior (which also receives the nodule I\textsubscript{1}), learn to map the input into a low dimensional latent space which encodes the distribution of all possible segmentation variants for the given input. In particular, we configured a latent vector of 6 features (or dimensions) as in the original paper, Figure-\ref{fig:arch_all}c. This network was trained to maximize the ELBO function composed by the pixel-wise binary cross entropy between the predicted and the ground truth segmentation, and the KL-divergence between the posterior (which can see the future image of the nodule) and the prior distributions. By sampling on the latent features of the prior network, this method allows generating multiple segmentations at inference time. 

The last generative approach consisted on a conditional GAN named Pix2Pix \citep{isola2017image}. The framework allows learning, in a model-free fashion, a mapping between two images. In our case the two images were a tumour image I\textsubscript{0} and a segmentation image Y at T\textsubscript{1}. The proposed network (P2P$\_$GAN) is composed by two networks; a generator formed by U-ResNet configured with dropout (p=0.5) along the decoder (no specific locations were indicated by the authors), and a discriminator composed by the encoder part of a U-ResNet, Figure-\ref{fig:arch_all}d. These two networks learn to generate images that are as similar as real ones, as well as to discriminate between images that are increasingly similar between real and fake ones. This network was trained as suggested by the authors using the $\Lb \textsubscript{cGAN}$ loss: 
\begin{center}
$
\Lb \textsubscript{cGAN}(G,D) = \mathop{\mathbb{E}}\textsubscript{I\textsubscript{1},Y} [log   D(I\textsubscript{1},Y)] +
\mathop{\mathbb{E}}\textsubscript{I\textsubscript{0},I\textsubscript{1},z} [log (1-D(I\textsubscript{1},G(I\textsubscript{0},z)))],
$
\end{center}

which represents the sum between the discriminator $D$ loss (i.e. binary cross entropy) of a nodule I\textsubscript{1} and the segmentation ground truth $Y$, and one minus the discriminator loss of a nodule I\textsubscript{1} and the segmentation $Y'$ produced by the generator at T\textsubscript{1}, i.e. $Y'=G(I\textsubscript{0},z)$. Additionally, a second term was added into this loss to figure out the fidelity of the generated samples with the ground truth. Thus, the L1 distance was computed between the generated sample $Y'$ and the truth $Y$. The final loss $G\textsuperscript{*}$ is as follows:
\begin{gather*}
G\textsuperscript{*} = arg\, \underset{G}{min}\, \underset{D}{max}\, \Lb \textsubscript{cGAN}(G,D) + \lambda \Lb \textsubscript{L1}(G). 
\end{gather*}
This approach allowed generating multiple samples by adding noise in the form of dropout, applied during training and testing time.

\subsection{Tumour growth assessment}



We assess tumour growth in terms of growth prediction, size and segmentation mask. For this, we considered not only the generative nature of our approach but also the fact of having per each nodule the opinion of up to three different radiologists. In the following, we provide further details regarding the tumour growth assessment.

\subsubsection{Metrics}


We proposed to evaluate how well the distributions produced by the generative model and the given ground-truth distributions agree. To this purpose, we considered two evaluation scenarios:

1) Using the expected value of the distributions, we computed conventional metrics such as
precision (Prec), recall (Rec), specificity (Spec) and balanced accuracy (Bacc) for growth prediction, mean absolute error (MAE) and mean squared error (MSE) for nodule growth, and Dice for segmentation fidelity.

2) Using confidence intervals, we defined the following metrics:
\begin{itemize}
\item For nodule growth prediction:

We proposed the metric Bacc$\_$2std. This computes the balanced accuracy between the radiologist tumour growth predictions (i.e. 1 if the tumour growth size was above 2 mm) and the predicted tumour growths at 2 standard deviations away from the estimated growth size means.  

To do this, we re-defined a true positive case as when the ground truth and the lower value of the predicted interval were above 2 mm. A false negative was when the ground truth was above 2 mm but the lower value of the interval was not. A true negative was when the ground truth and the upper value of the interval were less or equal to 2 mm. False positive was when the ground truth was less or equal to 2 mm but the upper value of the interval was not. 

\item For nodule growth size:

We proposed the ratio P(RX$\in$2std). This reports the proportion of tumours (over all tumours), whose growth size is within the interval (2 standard deviations away from the estimated tumour growth mean). 

To do this, we compared for each tumour, if the distance between the tumour growth size (ground truth) and the predicted growth size distribution was below the distance between the estimated mean with 2 standard deviations and the predicted growth size distribution. To compute this distance we used the Mahalanobis distance $D_{MH}$, which is the distance of a test point $x$, from the center of mass $m$, divided by the width of the ellipsoid defined by the covariance matrix $C$ in the direction of the test point.
\begin{gather*}
D_{MH}^2 = (x-m)\textsuperscript{T}C\textsuperscript{-1}(x-m). 
\end{gather*}

\item For nodule segmentation: 

We used the estimation of the Generalized Energy Distance (GED) \citep{szekely2013energy} metric. This metric reports the segmentation performance in terms of the variability in the ground truth as in the generated samples of the network.
\begin{center}
$
D_{GED}^2 = \dfrac{2}{n m} \sum\limits_{i=1}^{n}\sum\limits_{j=1}^{m} d(Y'_i,Y_j) -
\dfrac{1}{n^2} \sum\limits_{i=1}^{n}\sum\limits_{j=1}^{n} d(Y'_i,Y'_j) - 
\dfrac{1}{m^2} \sum\limits_{i=1}^{m}\sum\limits_{j=1}^{m} d(Y_i,Y_j). 
$
\end{center}

where $m$ and $n$ are the number of generated and ground truths segmentations, $Y'_i$ and $Y_j$ are a predicted and ground truth tumour segmentation and $d$ is the distance obtained using the 1-IoU metric. The resulting GED distance will be better the closer to 0.
\end{itemize}

\subsubsection{Ground truths}

To evaluate the models we used the annotations provided by the 3 different radiologists (RX0, RX1 and RX2). However, given that the annotations of the different radiologists may diverge (although all of them may still be correct), we derived two more ground truths, precisely, the mean of the radiologists annotations (RX$\_$mean), and the radiologist annotations that stands closest to our predictions (RX$\_$closest). Although the former is direct to obtain, the second has some particularities:
\begin{itemize}
\item For the deterministic model, we computed the closest radiologist tumour growth size and tumour growth prediction selecting the radiologist annotation with the minimum growth size difference with respect to the prediction. We computed the closest radiologist segmentation selecting the segmentation with highest Dice with respect to our prediction.
\item For the generative models, we computed the closest radiologist tumour growth size and tumour growth prediction selecting the radiologist annotation with the minimum Mahalanobis distance between the radiologist growth size and the estimated output distribution. We computed the closest radiologist segmentation selecting the radiologist segmentation with highest average Dice score obtained from the generated samples of the network.
\end{itemize}

\section{Experiments and Results}

\subsection{VH-Lung}

In this study we used a longitudinal lung CT dataset \citep{rafael2020re} for the follow-up analysis of incidental pulmonary nodules. In total, the cohort contains 161 patients (10 more cases compared to the previous version) with two thoracic CT scans per patient. The most relevant pulmonary nodule in each patient was located in each study by two different specialists. We address the reader to the source article for further details regarding the ethics, inclusion criteria, and acquisition protocol of the dataset. 

A new feature included in this updated version of the cohort is that up to 3 different clinicians ($RX0$, $RX1$ and $RX2$) reported the diameter size of the nodules ($D0$, $D1$) at the two different time-point ($T\textsubscript{0}$, $T\textsubscript{1}$) studies. From here, we computed the tumour growth by subtracting the diameters ($D1-D0$). The tumour growth mean in the dataset was 2.52$\pm$3.85 mm for RX0, 2.76$\pm$3.63 mm for RX1 and 2.68$\pm$4.01 mm  for RX2. The inter-observer mean absolute difference was 1.55 mm, whereas the inter-observer mean standard deviation was 0.97 mm (both metrics were computed pair-wise) \citep{popovic2017assessing}. The time interval between current and previous CT studies ranged from 32 to 2464 days. 

Tumour segmentations were obtained in a semi-automatic way, being visually verified and curated with the annotations provided by each of the radiologists (that is, location of the centroid, diameter and growth of the tumours) with a residual margin of 0.25 mm. 

\begin{figure}[!ht]
\centering
\includegraphics[scale=.4]{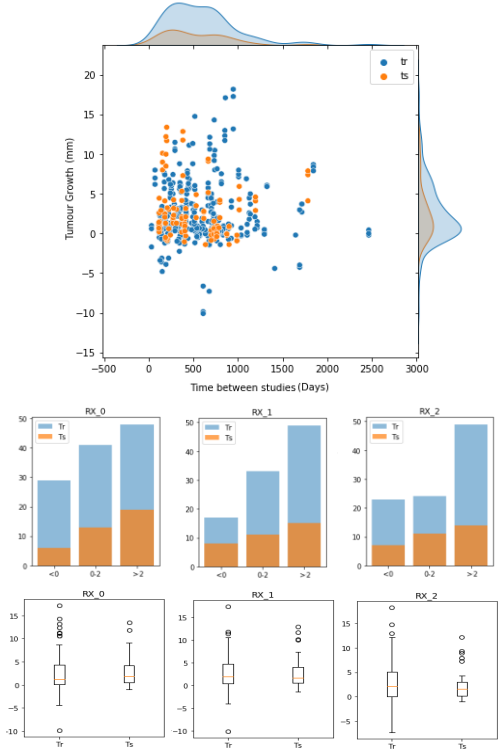}
\caption{Description of the training and test set cohorts in terms of tumour growth size (in mm), time between studies (days) and  the annotators (RX-0,1,2).}
\label{fig:cohort}%
\end{figure}

To train and evaluate the proposed methods, the whole data was randomly divided into training (70\%) and test (30\%) sets. In this process we assured that all entrances of the same nodule were in the same set in order to avoid data leakage between partitions. Therefore, the training set was composed of 313 (122 unique) nodule growth annotations from up to three different radiologists (118 for RX0, 99 for RX1 and 96 for RX2), whereas the test set was formed by 104 (39 unique) nodule growth annotations (38 for RX0, 34 for RX1 and 32 for RX2). Hence, for each data entrance (i.e. nodule growth annotation) of these partitions we had 2 nodule images (at T\textsubscript{0} and T\textsubscript{1}), 2 nodule segmentations (at T\textsubscript{0} and T\textsubscript{1}), a growth label (indicating whether it grew (1) or not (0)) and a growth size (in mm) corresponding to a particular radiologist. Further details regarding training and test set partitions can be seen in Figure-\ref{fig:cohort}. 

\subsection{Qualitative results}

\begin{figure*}[!ht]
\centering
\includegraphics[width=\textwidth]{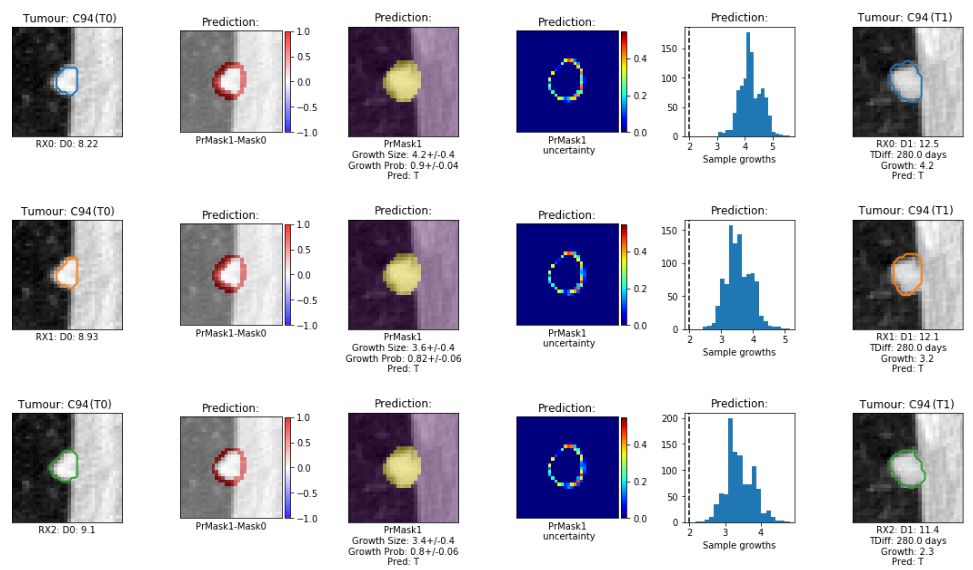}
\caption{Comparison of ground truth annotations and predictions from the U-HPNet for the tumour case C94.}
\label{fig:c94}
\end{figure*}

\begin{figure*}[!ht]
\centering
\includegraphics[width=\textwidth]{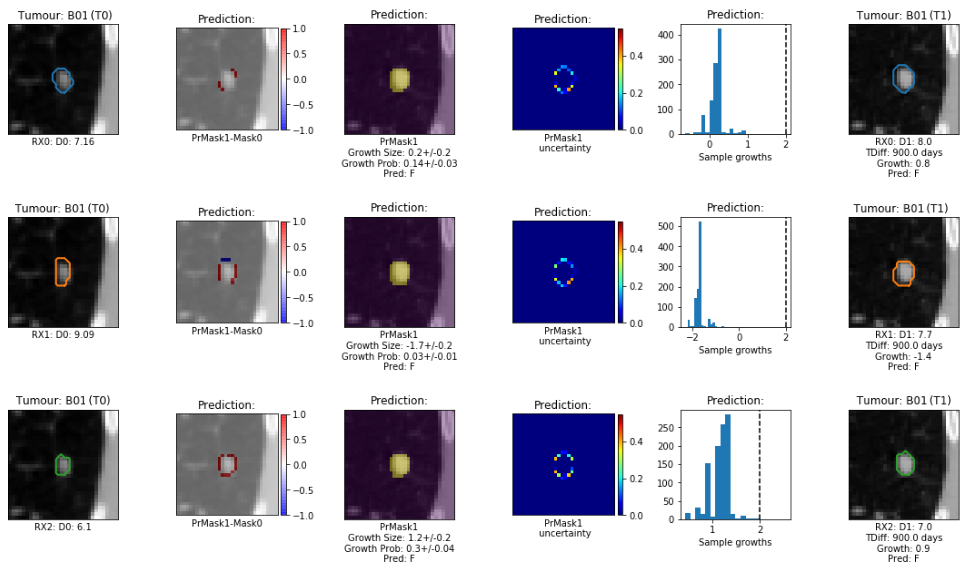}
\caption{Comparison of ground truth annotations and predictions from the U-HPNet for the tumour case B01.}
\label{fig:b01}
\end{figure*}

\begin{figure*}[!ht]
\centering
\includegraphics[width=\textwidth]{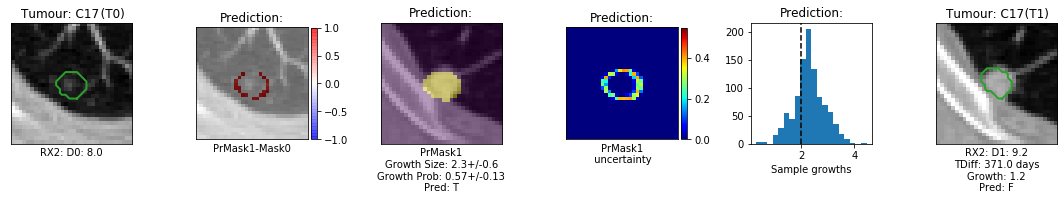}
\centering
\includegraphics[width=\textwidth]{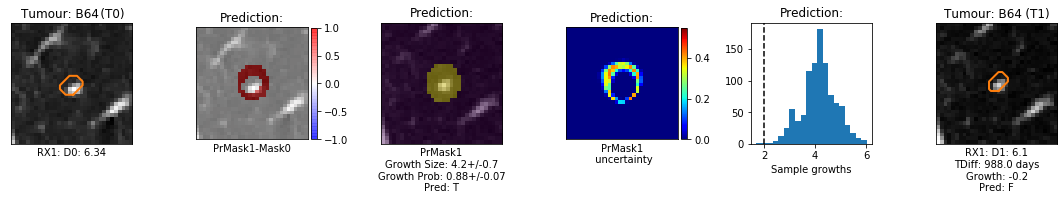}
\centering
\includegraphics[width=\textwidth]{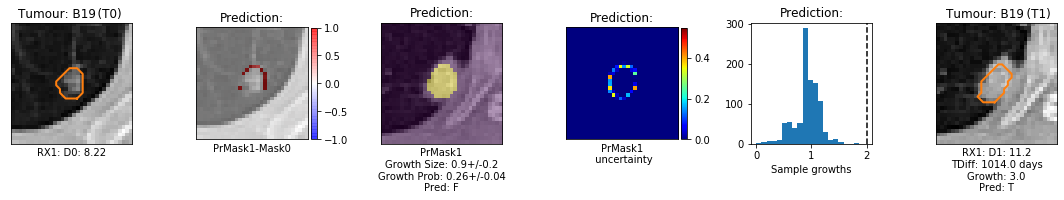}
\centering
\includegraphics[width=\textwidth]{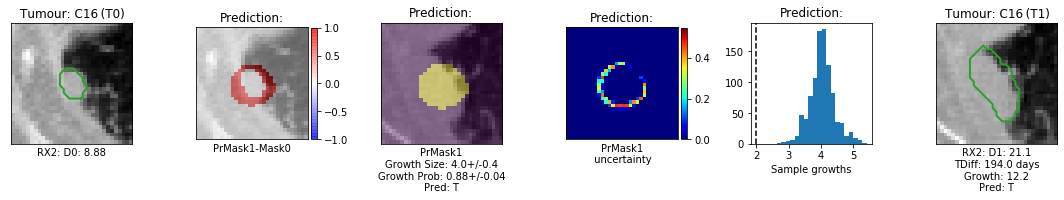}
\caption{Comparison of ground truth annotations and predictions from the U-HPNet for the tumours C17, B64, B19 and C16. In this example the network incorrectly predicts growth for the three first cases while for the last one it guesses the prediction although the predicted growth size is far from the radiologist measurement.}
\label{fig:b19}
\end{figure*}

We present some qualitative results (Figures-\ref{fig:c94},\ref{fig:b01},\ref{fig:b19}) obtained from the U-HPNet using lung nodules from the test set. We recall that these results were obtained (see section 3.2.3) from a post-processing aimed at obtaining the estimated growth probability, size and visualization together with its associated uncertainty.


The figures are composed of six columns, the first one shows the nodule at T\textsubscript{0} overlapped with the segmentation of a radiologist. The second column shows, overlapped with the nodule image at T\textsubscript{0}, the difference between the ground truth segmentation at T\textsubscript{0} and the estimated mean segmentation at T\textsubscript{1}. The third column provides, overlapped with the nodule image at T\textsubscript{1}, the estimated tumour mean segmentation. The fourth column visualizes the estimated uncertainty probability map with the per pixel-standard deviation. The fifth column shows the histogram of the (K=1000) estimated tumour growths (i.e. predicted diameter at T\textsubscript{1} minus radiologist diameter at T\textsubscript{0}). The last column shows the nodule image at T\textsubscript{1} overlapped with the segmentation of the radiologist at T\textsubscript{1}. 

The first tumour case (Figure-\ref{fig:c94}) was cancerous with 280 days between image studies. The tumour, attached to the chest wall, was annotated by three radiologists. Our method correctly predicted the existence of growth ($>$ 2mm) for each of the three radiologists, reporting high growth probabilities or estimated means (0.9, 0.82 and 0.8) and low uncertainties (0.04, 0.06 and 0.04) or standard deviations. The predicted tumour growth sizes were especially close (4.2, 3.6 and 3.4 mm) to those reported by the first two radiologists (4.2, 3.2 and 2.3 mm). 

The second tumour case (Figure-\ref{fig:b01}) was benign with almost 3 years between studies (900 days). All three radiologists did not detect any relevant tumour growth (diameter difference $\leq$ 2 mm) for this case. Our method correctly predicted the existence of no growth for each of the three radiologists, reporting low tumour growth probabilities (0.14, 0.03 and 0.3), especially for the first two radiologists, with low uncertainties (0.03, 0.01 and 0.05). The predicted tumour growth sizes were approximately less than 0.6 mm (0.2, -1.7 and 1.2 mm) apart from those reported by the radiologists (0.8, -1.4 and 0.9 mm). Moreover, the network agrees with the second radiologist to correctly predict tumour recession and also correctly guess to provide slightly higher probability, and greater tumour growth size for the third radiologist than for the other two.


Figure-\ref{fig:b19} shows 4 tumour examples in which our model struggles to find correct predictions. In particular for the tumour case C17, the model predicted a tumour growth size of 2.3$\pm$0.6 mm whereas the radiologist reported less than 2 mm (i.e. 1.2 mm). Nevertheless, the network provides a tumour growth probability close to 0.5 with an uncertainty of 0.14, which implies that the network is not highly confident on the nodule predictions. Looking at the ground truth provided for this case (first column of the figure), this mistake could be due to a probable overestimation of the diameter size of the nodule at T\textsubscript{0}. For the second tumour case, B64, the model predicted a growth size of 4.2$\pm$0.7 mm whereas the radiologist detected tumour recession (i.e. -0.2 mm). In this case, the network incorrectly provides high tumour growth probability and low uncertainty. However, if we look at the estimated per-pixel uncertainty image, the model correctly outputs a relevant quantity of uncertainty surrounding the nodule. In the third tumour case, B19, the model predicted a tumour growth size of 0.9$\pm$0.2 mm whereas the radiologist found a tumour growth of 3.0 mm. This could be caused by a relevant change in the context of the nodule, i.e. the nodule at T\textsubscript{1} was attached to the lung wall whereas at T\textsubscript{0} it was aerial. In the last case, C16, the model correctly predicted tumour growth. However, the radiologist indicated a high tumour increase of 12.2 mm whereas the network only detected 4.0$\pm$0.4 mm. If we observe the estimated segmentation mean image of the nodule, we see that the network missed to detect the longitudinal growth direction of the tumour due to not enough representation of this kind of tumour growth behaviour in the training set.

\subsection{Quantitative results}



Next, we detail the test performances of the proposed networks in terms of estimated nodule growth size, prediction and segmentation fidelity (for further details on the used metrics see section 3.4).
To obtain these performances, the methods were optimized with the training data using a 5-fold cross-validation, and tested with the testing set. The setup of the learning hyperparameters was the same for all methods, thus we used 1e-4 of learning rate, 8 of batch size, 200 epochs and Adam \citep{kingma2014adam} as optimization algorithm. 

The performances provided in the tables of the following sections show for each of the metrics, the mean and the standard deviation obtained from a bootstrapping process (with N=1000 iterations), in which for every iteration we performed a resample with replacement from the test set (N=104). 
\subsubsection{Nodule growth prediction}

The performances of the U-HPNet regarding lung nodule growth prediction are shown in Table-\ref{tab:growhpu1a}. These results were obtained using the annotations from three different radiologists (RX0, RX1 and RX2), their mean and the radiologists' annotations closest to our predictions (closest), with the intention to provide a more complete analysis of the performance of our method and to detect disparities between radiologists annotations (see section 3.4.2 for further details).

\begin{table}[!h]
\centering
\small\addtolength{\tabcolsep}{-2pt}
\begin{tabular}{l|cccc}
\hline
\hline
      RX & Bacc & Prec & Rec & Spec \\
\hline
RX0 &       0.49$\pm$0.07 &  0.46$\pm$0.11 &  0.42$\pm$0.11 &  0.56$\pm$0.11 \\
RX1 &       0.68$\pm$0.08 &  0.63$\pm$0.13 &  0.64$\pm$0.13 &  0.72$\pm$0.10 \\
RX2 &       0.67$\pm$0.09 &  0.58$\pm$0.15 &  0.59$\pm$0.15 &  0.75$\pm$0.10 \\
Mean &  0.55$\pm$0.04 &  0.50$\pm$0.07 &  0.47$\pm$0.07 &  0.63$\pm$0.06 \\
\textbf{Closest} &  \textbf{0.74}$\pm$\textbf{0.07} &  \textbf{0.65}$\pm$\textbf{0.12} &  \textbf{0.71}$\pm$\textbf{0.12} &  \textbf{0.76}$\pm$\textbf{0.08} \\
\hline
\hline
\end{tabular}
\caption{Nodule growth prediction performances obtained by the U-HPNet using expected means.}
\label{tab:growhpu1a}
\end{table}

Despite the heterogeneity in the morphology, density and location of the nodules, the fact of using a single image (I\textsubscript{0}) of the tumour and the variety on the time at which to make the predictions, our method was able to satisfactorily report positive performances, such as 0.74 of balanced accuracy, 0.71 of recall and 0.76 of specificity. 
 
\begin{figure}[!ht]
\centering
\includegraphics[scale=.45]{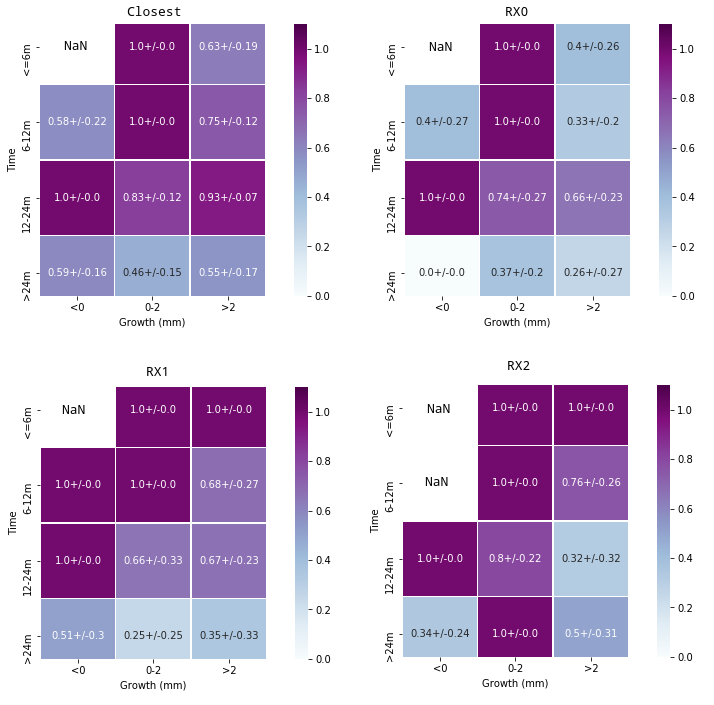}
\caption{Tumour growth prediction accuracy of the U-HPNet stratified by time to predict, nodule growth size and ground truth (i.e. Closest, RX0, RX1 and RX2).}
\label{fig:predTimeGrowth}%
\end{figure}

Further details regarding the tumour growth prediction of the U-HPNet are provided in Figure-\ref{fig:predTimeGrowth}. This figure shows the prediction accuracy stratified by the time to predict, the real growth of the nodules and the ground truth type. From this figure, we can observe that, the best performances were usually obtained when predicting in the range of 12 to 24 months, whereas the worst performances were obtained when predicting above 24 months. Also, the best performances were usually obtained when the nodules had a growth between 0 and 2 mm. 

Additionally, we provide Table-\ref{tab:growhpu1b} with the balanced accuracy obtained at 2 standard deviations away from the expected means.

\begin{table}[!h]
\centering
\begin{tabular}{l|cccc}
\hline
\hline
      RX & Bacc$\_$2std \\
\hline
 RX0 &  0.37$\pm$0.08 \\
 RX1 &  0.53$\pm$0.09 \\
 RX2 &  0.57$\pm$0.10 \\
 Mean & 0.45$\pm$0.05 \\
 \textbf{Closest} &  \textbf{0.57}$\pm$\textbf{0.08} \\
\hline
\hline
\end{tabular}
\caption{Nodule growth balanced accuracy performances for the U-HPNet obtained within 2 standard deviations from the mean.}
\label{tab:growhpu1b}
\end{table}

From this table, we observe that the performances are clearly below the 0.74 of balanced accuracy obtained using single point estimates (i.e. the expected mean). The reason is because Bacc$\_$2std reports the balanced accuracy at 2 standard deviations away from the expected mean. Therefore, at this extreme our approach is still able to correctly predict tumour growth in 57\% of the test cases. 

\subsubsection{Nodule growth size}

We also computed the nodule growth size performance of the U-HPNet.
Table-\ref{tab:sizehpu1} provides the mean absolute error (MAE) and mean squared error (MSE), obtained from the comparison of the estimated tumour growth size mean of the network for the different ground truths. Also, this table reports the metrics regarding the probability of finding the radiologist measurements within 2 standard deviation P(RX$\in$2std) from the estimated tumour growth size mean.  

\begin{table}[!h]
\centering
\begin{tabular}{l|cc|cc}
\hline
\hline
      RX &              MAE $\downarrow$&               MSE $\downarrow$&    P(RX$\in$2std) \\
\hline
RX0 &  2.99$\pm$0.42 &  16.60$\pm$4.94 &  0.20$\pm$0.06 \\
RX1 &  2.52$\pm$0.45 &   13.11$\pm$4.40 &  0.21$\pm$0.07 \\
RX2 &  2.62$\pm$0.41 &  11.86$\pm$2.98 &  0.28$\pm$0.08 \\
Mean &  2.83$\pm$0.23 &  14.40$\pm$2.39 &  0.17$\pm$0.04 \\
\textbf{Closest} &  \textbf{1.74}$\pm$\textbf{0.34} &   \textbf{7.55}$\pm$\textbf{2.87} &  \textbf{0.44}$\pm$\textbf{0.08} \\
\hline
\hline
\end{tabular}
\caption{Nodule growth size performances of the U-HPNet using the estimated mean (MAE, MSE) and the interval composed by the mean and 2 standard deviations. A down arrow next to a metric means that the metric is more accurate the smaller the value.}
\label{tab:sizehpu1}
\end{table}

As we observe from Table-\ref{tab:sizehpu1} the best performances on tumour growth size reported a MAE of 1.74 mm close to the 1.55 mm of inter-observer mean absolute difference. 
Moreover, this method reported that in 44\% of the cases, the exact tumour growths annotated by the radiologists were found at 2 standard deviations from the mean. Further details are exposed in the discussion section.

\subsubsection{Nodule segmentation}

Subsequently, we report the performance of the U-HPNet for predicting accurate future nodule segmentations. To do this, we computed for each nodule of the test set, the average Dice score obtained from each generated tumour segmentation with respect to the proposed radiologist segmentation. Table-\ref{tab:diceMean} summarizes the resulting Dice performances for each of the radiologists. 

\begin{table}[!h]
\centering
\small\addtolength{\tabcolsep}{-2pt}
\begin{tabular}{ccccc}
\hline
\hline
RX0 &              RX1 &              RX2 &             Mean  & Closest \\
\hline
0.74$\pm$0.02 &  0.77$\pm$0.02 &  0.75$\pm$0.02 & 0.76$\pm$0.02 &  \textbf{0.78}$\pm$\textbf{0.02} \\
\hline
\hline
\end{tabular}
\caption{Nodule segmentation performances of the U-HPNet for each of the ground truths.}
\label{tab:diceMean}
\end{table}

The best Dice score was 78\%, achieved for the closest radiologists ground truths. Complementary, we also computed the GED metric to report the ability of the network to generate accurate and diverse future tumour segmentations (Table-\ref{tab:gedhpu1}). From this result, we remark that a high segmentation agreement (i.e. 0.14 of 1-IoU) was found between ground truths (YY). This may explain why an small variability (i.e. 0.04 of 1-IoU) was also found between predicted segmentations (Y'Y').

\begin{table}[!h]
\centering
\begin{tabular}{l|ccc}
\hline
\hline
         GED &             2*(Y'Y) &              Y'Y' &              YY \\
\hline
 0.29$\pm$0.04 &  0.48$\pm$0.04 &  0.04$\pm$0.01 &  0.14$\pm$0.01 \\
\hline
\hline
\end{tabular}
\caption{GED nodule segmentation performance of the U-HPNet. Each score reports 1-IoU metric.}
\label{tab:gedhpu1}
\end{table}

As a general observation about the results shown previously (Tables-\ref{tab:growhpu1a},\ref{tab:growhpu1b},\ref{tab:sizehpu1} and \ref{tab:diceMean}), we should mention that the best performances were always achieved with the radiologists' annotations closest to our predictions. This ground truth is as important as any of the others since, as mentioned in section 3.4.2, the annotations of each of the radiologists were assumed to be equally valid. Therefore, we already expected that our method would work somewhat better with this criterion since, by definition, for each tumor growth prediction we compared it with the radiologist's annotation closest to that prediction. Regarding the rest of the ground truths, the results obtained with the RX1 and RX2 annotations were found near to the best performances,  while the results obtained with RX0 annotations and the mean of the radiologists' annotations were the lowest.

\subsection{Ablation studies}

An ablation study was made to isolate the effects of the different components of the U-HPNet using the radiologists' annotations closest to our predictions. Table-\ref{tab:growthMeta} shows the different network setups evaluated and their acronym for better identification.

\begin{table}[!h]
\centering
\begin{tabular}{l|cccc}
\hline
\hline

   Acronym    &              $\Lb \textsubscript{rec}$ &  Attention &  D0 \\

\hline
BD0               & BCE    & \xmark     & \checkmark\\
ID0                & IoU    & \xmark     & \checkmark\\
IDD0               & IoU+L1 & \xmark     & \checkmark\\
IDAOD0              & IoU+L1 & \checkmark & \xmark\\
\hline
U-HPNet              & IoU+L1 & \checkmark & \checkmark\\

\hline
\hline
\end{tabular}
\caption{Different network setups of the U-HPNet configured in the ablation study.}
\label{tab:growthMeta}
\end{table}

\begin{table}[!h]
\centering
\small\addtolength{\tabcolsep}{-2pt}
\begin{tabular}{l|cccc}
\hline
\hline

       &              Prediction  &               Size $\downarrow$&      Segmentation  \\
       & (Bacc) & (MAE)& (Dice) \\

\hline
BD0         & 0.66$\pm$0.09  &  1.93$\pm$0.35 & 0.74$\pm$0.02\\
ID0         & 0.72$\pm$0.08  &  1.80$\pm$0.35 & 0.79$\pm$0.02\\
IDD0        & \textbf{0.75}$\pm$\textbf{0.08}  &  1.84$\pm$0.37 & 0.80$\pm$0.02\\
IDAOD0       & 0.74$\pm$0.08  &  1.79$\pm$0.38 & \textbf{0.81}$\pm$\textbf{0.02}\\
\hline
U-HPNet           &0.74$\pm$0.08 & \textbf{1.74}$\pm$\textbf{0.34} & 0.78$\pm$0.02\\

\hline
\hline
\end{tabular}
\caption{Performance comparison between the different U-HPNet setups.}
\label{tab:growthMeanConfigs}
\end{table}

Table-\ref{tab:growthMeanConfigs} shows the performances obtained for tumour growth prediction, size and segmentation for each of the network setups using their estimated means. An interesting observation to note from these results is that the configurations using IoU clearly outperformed the setup using BCE (BD0). Particularly, a rise of 0.09 in Bacc was achieved with the IDD0, an improvement of 0.19 mm in MAE was obtained with the U-HPNet and an increase of 0.07 in Dice score was reached with the IDAOD0. 
Also, the networks using attention (i.e. U-HPNet and IDAOD0) obtained the best performances in terms of MAE, in particular the U-HPNet obtained the lowest value with 1.73 mm. Regarding the Dice score all networks using IoU loss obtained performances above 0.78, although the IDAOD0 with 0.81 was the one with the highest performance. 

\begin{table}[!h]
\centering
\small\addtolength{\tabcolsep}{-2pt}
\begin{tabular}{l|cccc}
\hline
\hline

       &              Prediction  &    Size &      Segmentation  $\downarrow$\\
       &             (Bacc$\_$2std) &    (P(RX$\in$2std)) &      (GED) \\

\hline
BD0   & 0.08$\pm$0.04 & \textbf{0.87}$\pm$\textbf{0.05} & \textbf{0.24}$\pm$\textbf{0.02}\\
ID0   & 0.59$\pm$0.08 & 0.38$\pm$0.08 & 0.31$\pm$0.02\\
IDD0  & 0.64$\pm$0.08 & 0.33$\pm$0.08 & 0.31$\pm$0.03\\
IDAOD0 & \textbf{0.67}$\pm$\textbf{0.08} & 0.36$\pm$0.08 & 0.30$\pm$0.04\\
\hline
U-HPNet     &  0.57$\pm$0.08 & 0.44$\pm$0.08 & 0.29$\pm$0.04\\

\hline
\hline
\end{tabular}
\caption{Generative ability comparison between the different U-HPNet setups.}
\label{tab:growthInterConfigs}
\end{table}

Table-\ref{tab:growthInterConfigs} shows the performances of the generative ability of the different network configurations. The best option regarding prediction performance was IDAOD0 with 0.67 of Bacc$\_$2std. The best network for size and segmentation was BD0, although it reported an unacceptable prediction performance of 0.08 in Bacc$\_$2std due to a high variability in the generated samples. If we do not consider this option, the best option was the U-HPNet either in P(RX$\in$2std) and GED.

\subsection{Comparison with other networks}

We evaluated 4 different alternative deep networks for nodule growth estimation using the radiologists' annotations closest to our predictions, to enable their comparison with the proposed method. In particular, we evaluated 1 deterministic (U-Net) and 3 generative architectures (GAN-P2P, BAYES$\_$TDO, SPU). Table-\ref{tab:growhMeanCompSOTA} shows the performances obtained for these models regarding nodule growth prediction (Bacc), size (MAE) and segmentation quality (Dice) using the predicted value for the deterministic approach and, using the expected mean of the output distribution for the generative approaches. 

\begin{table}[!h]
\centering
\begin{tabular}{l|cccc}
\hline
\hline

       &              Prediction  &     Size $\downarrow$&      Segmentation  \\
       & (Bacc) & (MAE)& (Dice) \\

\hline
U-Net      &  0.64$\pm$0.09 & 2.94$\pm$0.43 & 0.77$\pm$0.02 \\
BAYES$\_$TD  &  0.67$\pm$0.08 & 2.29$\pm$0.45 & \textbf{0.78}$\pm$\textbf{0.02} \\
SPU        &  0.73$\pm$0.08 & 2.14$\pm$0.46 & 0.77$\pm$0.01 \\
P2P$\_$GAN   &  0.69$\pm$0.07 & 2.62$\pm$0.43 & 0.71$\pm$0.02 \\
\hline
U-HPNet    & \textbf{0.74}$\pm$\textbf{0.08} & \textbf{1.74}$\pm$\textbf{0.34} & \textbf{0.78}$\pm$\textbf{0.02}\\

\hline
\hline
\end{tabular}
\caption{Performance comparison with alternative networks for tumour growth using the expected mean.}
\label{tab:growhMeanCompSOTA}
\end{table}

From the four alternative methods, the SPU obtained the best Bacc score with 0.73 and MAE with 2.14 mm. In contrast, the BAYES$\_$TD method obtained the best Dice score with 0.78. If we compare these results with the U-HPNet none of them could outperform their results neither in terms of prediction, size or segmentation.

In Table-\ref{tab:growhInterCompSOTA}, we summarize the performances regarding the generative ability to report accurate results. In particular, we provide nodule growth prediction using BA$\_$2std metric, nodule size using P(RX$\in$2std) and estimated nodule segmentation using GED.

\begin{table}[!h]
\centering
\begin{tabular}{l|c|c|c}
\hline
\hline

       &              Prediction  &    Size $\downarrow$ &      Segmentation \\
       &             (Bacc$\_$2std) &    (P(RX$\in$2std)) &      (GED)   \\
\hline
BAYES$\_$TD  & 0.46$\pm$0.08  & 0.49$\pm$0.08 & 0.27$\pm$0.03\\
SPU        & 0.28$\pm$0.06  & \textbf{0.68}$\pm$\textbf{0.08} & \textbf{0.23}$\pm$\textbf{0.02}\\ 
GAN-P2P    & 0.26$\pm$0.07  & 0.67$\pm$0.08 & 0.25$\pm$0.04\\
\hline
U-HPNet    & \textbf{0.57}$\pm$\textbf{0.08} & 0.44$\pm$0.08 & 0.29$\pm$0.04 \\
\hline
\hline
\end{tabular}
\caption{Generative performance of alternative networks for tumour growth.}
\label{tab:growhInterCompSOTA}
\end{table}

The best Bacc$\_$2std score was 0.46 for the BAYES$\_$TD, the best P(RX$\in2std$) was for the SPU with 0.68 and the best GED with 0.25 mm for the GAN-P2P. If we compare these results with the U-HPNet, we observe that other methods showed better segmentation and size generative ability to capture the ground truth, however this made them to be less accurate with lowest prediction performances.

\section{Discussion}


With the aim of supporting radiologists in the early detection of lung cancer, we proposed a new predictive method capable of estimating tumour growth at a given time. In line with current clinical practice, our method predicts tumour progression when there is substantial growth (i.e., more than 2 mm) in the longest diameter of the pulmonary nodule \citep{macmahon2017guidelines}. Although this criterion is commonly used for its simplicity and applicability, it entails significant inter-observer \citep{han2018influence} variability that may impact on the reliability of the predictive models. Along with the inter-observer variability, other inter-related factors may also have a direct impact on the trustworthiness of the estimator, such as the ambiguity, partiality or scarcity of the data to model. Therefore, in medical settings it is important that predictive models also provide a measure of uncertainty, which is especially of interest when complex or doubtful cases have to be assessed. 
 
In this work we have taken this aspect into account and we have built a predictive model capable of also estimating the associated uncertainty when predicting tumour growth. To do this, we collected a longitudinal dataset with more than 160 selected pulmonary tumours with two CT images per case (taken at different time-points), labelled by up to three different radiologists. To model these data, we opted for a generative deep learning approach as opposed to the deterministic approaches used to date  \citep{li2020learning, wang2019toward} for lung tumour growth prediction. The suitability of the generative approach was already proved in \cite{petersen2019deep}, where they modeled glioma tumour growth using an early probabilistic and generative framework \citep{kohl2018probabilistic} to estimate the tumour growth output distribution. Nonetheless, tumour growth prediction was not quantified, model uncertainty was not reported, and multiple observer variability was not addressed. 

To address the aforementioned aspects, we relied on a more recent hierarchical generative and probabilistic framework \citep{kohl2019hierarchical} to estimate the output distribution of the future lung tumour appearance (at T\textsubscript{1}) conditioned on the previous image of the nodule (at T\textsubscript{0}). Our method (U-HPNet) extended this framework with the following modifications. First, we used smaller image patches (32x32) to focus on the tumour and its immediate surrounding tissues, and to reduce the number of parameters to be adjusted by the network. Second, we added two new features to the network to extract additional patterns from the tumour images: the time to predict, and the diameter of the nodule (at T\textsubscript{0}). Third, we integrated an attention mechanism \citep{oktay2018attention} in the decoder part of the network to boost its performance. Fourth, we proposed a new reconstruction loss function composed of the IoU and the L1 distance to provide more accurate segmentation and diameter estimations. Finally, we created a new post-processing module that applies Monte-Carlo sampling to estimate the mean and standard deviation of the tumour growth prediction, diameter growth and segmentation of a given nodule at an specific time. 

We evaluated the U-HPNet using the annotations provided by 3 different radiologists, but also with their average and the radiologists' annotations closest to our estimates, to provide a more complete assessment of our approach and to detect possible divergences between the experts. 

Regarding the evaluation of our approach using the expected values, the best results were obtained using the radiologists' annotations closest to our predictions. This ground truth criterion always reports real radiological annotations (specifically, the closest ones to our predictions), therefore since we take all radiologists' opinions equally, in a sense, this criteria is equally comparable to any of the three radiological criteria available in the study. In particular, we achieved 74\% of tumour growth balanced accuracy (Bacc), 1.73 mm of diameter mean absolute error (MAE) and 78\% of Dice score (Tables-\ref{tab:growhpu1a}, \ref{tab:sizehpu1}, \ref{tab:diceMean}). Near to these results, we found the performances obtained with RX1 and RX2 annotations. Specifically, for RX1 we achieved 0.68 of Bacc, 2.52 of MAE and 0.77 of Dice score (Tables-\ref{tab:growhpu1a}, \ref{tab:sizehpu1}, \ref{tab:diceMean}). Lower performances were found using the RX0 annotations and the mean of all radiologists, especially on tumours with a growth size greater than 2 mm and predictions over 24 months (Figure-\ref{fig:predTimeGrowth}).

Compared to similar recent work in the literature \citep{li2020learning}, they reported higher balanced accuracy scores (86\%) but much lower segmentation Dice scores (64\%) than us. Results however are not fully comparable since both networks used different in-house cohorts, with different tumour case complexities, and both defined tumour progression differently, theirs relied on a tailored volumetric threshold and ours on the diameter growth convention established in radiological guidelines (\citep{macmahon2017guidelines}).

We also evaluated the ability of the network to produce consistent samples matching with the ground truths. To this end, we proposed different metrics (see section 3.4), i.e. the balanced accuracy for tumour growth prediction in an interval of 2 standard deviations (Bcc$\_$2std), the probability of matching with the tumour growth size in an interval of 2 standard deviations (P(RX$\in$2std)) and the generalized energy distance for tumour segmentation (GED). 
Our method achieved the best performances with the closest radiologists criterion, in particular 57\% of Bacc$\_$2std, and 44\% of P(RX$\in$2std) (Tables-\ref{tab:growhpu1b},\ref{tab:sizehpu1}). 
These values reflect that our approach still has room for improvement to make the estimated tumour growth sizes more accurate (i.e. bringing the tumour growth size mean closer to the radiologists ground truths). However, we should stress that these performances (as seen in Figure-\ref{fig:predTimeGrowth}) were affected especially by complex cases with higher uncertainty (i.e. with a temporal prediction distance above 24 months). Different solutions could be applied to improve these performances such as acquiring more tumour cases (e.g. especially on those cases where the method was not as accurate), using more aggressive data augmentation techniques (e.g. generating synthetic tumours); or using volume images, rather than single slices, to extract better predictive features. Breaking down the GED performance (Table-\ref{tab:gedhpu1}), we observed that the network obtained 23\% of segmentation variability between predicted and ground truths (Y'Y), being not far from 14\% of inter-observer variability (YY). Also, the network showed a relatively small variability of 4\% between the generated sample segmentations (Y'Y'). This may indicate that the network, during training, preferred to concentrate the predictions around the mean rather than predict highly disperse values in order to optimize performance. 


For a better understanding on the effects of the main components of the network, we provided an ablation study with different network configurations using the closest radiologists criterion (Table-\ref{tab:growthMeta}). From this analysis, we obtained that the largest improvement was achieved replacing binary cross entropy (BCE) by IoU in the reconstruction loss. This can be observed by comparing BD0 and ID0 networks. Specifically, the Bacc increased approximately 7\%, MAE decreased almost 0.2 mm, and the Dice improved to nearly 5\%. Moreover, the Bacc$\_$2std raised to almost 60\% (Table-\ref{tab:growthMeanConfigs}). Different reasons may explain the suitability of using IoU for this problem. First, this loss is robust to data unbalance. Second, IoU had values with similar magnitude to the KL-divergence distance, allowing a better optimization of the network than using BCE. Despite the benefits of using IoU, we realized that the P(RX$\in$2std) and GED decreased significantly due to higher variability around the estimated mean. A second network configuration (IDD0) allowed improving previous performance limitations. In particular, this network incorporated the L1 distance between the predicted and ground truth diameters in the reconstruction loss together with IoU. Results showed that the IDD0 network increased its growth prediction performance (3\% in Bacc and 5\% in Bacc$\_$2std) and segmentation ability (1\% in Dice and GED), despite slight decrease of performance in diameter growth prediction (0.05 mm in MAE and 4\% in P(RX$\in$2std)). Adding attention (current U-HPNet network) in the decoder part of the sub-networks, outperformed the IDD0, precisely, reducing 0.1 mm in MAE and increasing 10\% in P(RX$\in$2std). However, it implied a certain increase also in the estimated diameter growth variability, reducing 1\% its Bacc and 7\% of Bacc$\_$2std.  A final comparison was performed between IDAOD0 and U-HPNet to obtain the importance of adding nodule diameter (at T\textsubscript{0}) in the input of the U-HPNet. According to the results, using this feature we reduced 0.6 mm of MAE, increased 8\% the P(RX$\in$2std)), and consequently improved the Bacc almost 2\%. This reflects that using D0, the network was able to predict more accurately the diameter growth of the nodule. However, this feature increased the estimated diameter growth variability, resulting in 9\% decrease of Bacc$\_$2std and 2\% of Dice.

Due to the lack of similar studies for early lung tumour growth prediction, we built different alternative networks to allow their comparison. In particular, we proposed a deterministic (U-Net), and 3 different generative networks: Bayesian dropout (BAYES$\_$TD), probabilistic U-Net (SPU), Pix2Pix GAN (P2P$\_$GAN). The comparison was performed using the closest radiologists criterion. Results from Tables-\ref{tab:growhMeanCompSOTA},\ref{tab:growhInterCompSOTA} showed that, using the estimated sample means, the generative approaches outperformed the performances reported by the deterministic network (U-Net). This result consolidates the suitability of the generative approach for this type of problem. Also, among all generative methods, the U-HPNet obtained the best performance metrics using the estimated means (i.e. in tumour growth prediction, size and segmentation). Regarding the metrics measuring the generative ability of the networks, the U-HPNet obtained the best Bacc$\_$2std although the poorest P(RX$\in$2std). In contrast, the SPU reported a large sample variability in tumour growth size as shown by the highest performance in P(RX$\in$2std) and GED, but one of the lowest performances in Bacc$\_$2std. The GAN-P2P, similarly to the SPU, obtained high P(RX$\in$2std) and GED, but poor Bacc$\_$2std due to high variability in the sample distribution of tumour growth size. The BAYES$\_$TD in contrast obtained lower variability in tumour growth size achieving better P(RX$\in$2std) and GED than the U-HPNet but lower Bacc$\_$2std. Interestingly BAYES$\_$TD and U-HPNet networks reported rather similar generative performances, despite employing two different ways to generate samples (by weight randomization and by randomly selecting a vector in the latent space). Thus, combining both approaches could help to disentangle different types of uncertainties (as in \cite{hu2019supervised} for tumour segmentation) and disclose a potential increase in the performance of the network.

Our method still has a number of limitations.
First, the number of cases analysed in this study was low, which clearly impacted on the reported performances of our approach due to its data eager nature. However, this data was clinically validated, and selected by different radiologists according to their relevance and interest. 
Second, segmentations were generated semi-automatically according to the original diameter, growth and centroid annotations with a final visual expert validation. However, we believe using manual expert segmentations could make our method more precise, especially in the contour of the tumours. 
Third, our method relied on a single axial slice of the tumour to predict tumour growth. However, tumour growth is a tri-dimensional biological process, hence using volumetric images may allow capturing further relevant features and patterns to explain better the tumour progression. Nevertheless, using 2D information made our solution more compact, with less parameters to fit, faster to train and more suitable for smaller datasets.

Beyond this work, further efforts in fine-tuning the proposed approaches are required such as exploring different number of layers, latent hierarchies, loss weights factors and other optimization parameters. Also, future extensions are suggested along this article, such as exploring a 3D version of the network, deepen in the uncertainty ability of the network, evaluate its integration with Bayesian dropout or adversarial learning, and incorporate the newest advances in deep learning to extract better spatial and temporal features from the the tumours.

\section{Conclusion}

In this article, we addressed early lung tumour growth prediction as a multimodal output problem, as opposed to existing solutions that provide deterministic outputs. Several reasons motivated our decision such as the complexity of the problem, the inter-observer variability, or the importance of estimating uncertainty in medical settings. To this end, we adapted an existing deep hierarchical generative and probabilistic framework to encode the initial image of the nodules in a continuous multidimensional latent space, to sample from it, and to generate multiple consistent future tumour segmentations conditioned on the given nodule. 

Our network (U-HPNet) extended the original framework with the intention to predict and quantify tumour growth, as well as to visualize the future semantic appearance of the tumour. Therefore, we added new context features (i.e. the time to predict and the initial nodule diameter measured by the specialist), we used a new reconstruction loss (combining IoU and diameter distance), and we integrated an attention mechanism in the decoder parts of the network. Finally, we attached a new post-processing module on the network to perform Monte Carlo sampling, and retrieve the estimated tumour growth probability, size and segmentation, along with their associated uncertainty. 

The network was trained and evaluated on a longitudinal cohort with more than 160 cases. Best performances reported a tumour growth balanced accuracy of 74\%, a tumour growth size MAE of 1.77 mm and a tumour segmentation Dice score of 78\%. Finally, we compared the performance of our method with 4 different networks based in a U-Net, probabilistic U-Net, Bayesian dropout and Pix2Pix GAN. The U-HPNet outperformed the proposed alternatives for tumour growth prediction, size and segmentation.

\section*{Acknowledgments}
This work was partially funded by the Industrial Doctorates Program
(AGAUR) grant number DI087, and the Spanish Ministry of Economy and Competitiveness (Project INSPIRE FIS2017-89535-C2-2-R, Maria de Maeztu Units of Excellence Program MDM-2015-0502).

\bibliographystyle{model2-names.bst}\biboptions{authoryear}
\bibliography{references}

\begin{thebibliography}{61}
\expandafter\ifx\csname natexlab\endcsname\relax\def\natexlab#1{#1}\fi
\providecommand{\url}[1]{\texttt{#1}}
\providecommand{\href}[2]{#2}
\providecommand{\path}[1]{#1}
\providecommand{\DOIprefix}{doi:}
\providecommand{\ArXivprefix}{arXiv:}
\providecommand{\URLprefix}{URL: }
\providecommand{\Pubmedprefix}{pmid:}
\providecommand{\doi}[1]{\href{http://dx.doi.org/#1}{\path{#1}}}
\providecommand{\Pubmed}[1]{\href{pmid:#1}{\path{#1}}}
\providecommand{\bibinfo}[2]{#2}
\ifx\xfnm\relax \def\xfnm[#1]{\unskip,\space#1}\fi
\bibitem[{Ayhan and Berens(2018)}]{ayhan2018test}
\bibinfo{author}{Ayhan, M.S.}, \bibinfo{author}{Berens, P.},
  \bibinfo{year}{2018}.
\newblock \bibinfo{title}{Test-time data augmentation for estimation of
  heteroscedastic aleatoric uncertainty in deep neural networks}, in:
  \bibinfo{booktitle}{Medical {I}maging with {D}eep {L}earning}.
\newblock \URLprefix \url{https://openreview.net/forum?id=rJZz-knjz}.
\bibitem[{Bankier et~al.(2017)Bankier, MacMahon, Goo, Rubin, Schaefer-Prokop
  and Naidich}]{bankier2017recommendations}
\bibinfo{author}{Bankier, A.A.}, \bibinfo{author}{MacMahon, H.},
  \bibinfo{author}{Goo, J.M.}, \bibinfo{author}{Rubin, G.D.},
  \bibinfo{author}{Schaefer-Prokop, C.M.}, \bibinfo{author}{Naidich, D.P.},
  \bibinfo{year}{2017}.
\newblock \bibinfo{title}{Recommendations for measuring pulmonary nodules at
  {CT}: a statement from the {F}leischner society}.
\newblock \bibinfo{journal}{Radiology} \bibinfo{volume}{285},
  \bibinfo{pages}{584--600}.
\bibitem[{Basu et~al.(2019)Basu, Wagstyl, Zandifar, Collins, Romero and
  Precup}]{basu2019early}
\bibinfo{author}{Basu, S.}, \bibinfo{author}{Wagstyl, K.},
  \bibinfo{author}{Zandifar, A.}, \bibinfo{author}{Collins, L.},
  \bibinfo{author}{Romero, A.}, \bibinfo{author}{Precup, D.},
  \bibinfo{year}{2019}.
\newblock \bibinfo{title}{Early prediction of alzheimer’s disease progression
  using variational autoencoders}, in: \bibinfo{booktitle}{International
  {C}onference on {M}edical {I}mage {C}omputing and {C}omputer-{A}ssisted
  {I}ntervention}, \bibinfo{organization}{Springer}. pp.
  \bibinfo{pages}{205--213}.
\bibitem[{Baumgartner et~al.(2019)Baumgartner, Tezcan, Chaitanya, H{\"o}tker,
  Muehlematter, Schawkat, Becker, Donati and Konukoglu}]{baumgartner2019phiseg}
\bibinfo{author}{Baumgartner, C.F.}, \bibinfo{author}{Tezcan, K.C.},
  \bibinfo{author}{Chaitanya, K.}, \bibinfo{author}{H{\"o}tker, A.M.},
  \bibinfo{author}{Muehlematter, U.J.}, \bibinfo{author}{Schawkat, K.},
  \bibinfo{author}{Becker, A.S.}, \bibinfo{author}{Donati, O.},
  \bibinfo{author}{Konukoglu, E.}, \bibinfo{year}{2019}.
\newblock \bibinfo{title}{Phiseg: Capturing uncertainty in medical image
  segmentation}, in: \bibinfo{booktitle}{International {C}onference on
  {M}edical {I}mage {C}omputing and {C}omputer-{A}ssisted {I}ntervention},
  \bibinfo{organization}{Springer}. pp. \bibinfo{pages}{119--127}.
\bibitem[{Bengio et~al.(2017)Bengio, Goodfellow and Courville}]{bengio2017deep}
\bibinfo{author}{Bengio, Y.}, \bibinfo{author}{Goodfellow, I.},
  \bibinfo{author}{Courville, A.}, \bibinfo{year}{2017}.
\newblock \bibinfo{title}{Deep learning}. volume~\bibinfo{volume}{1}.
\newblock \bibinfo{publisher}{MIT press Massachusetts, USA:}.
\bibitem[{Ciompi et~al.(2017)Ciompi, Chung, Van~Riel, Setio, Gerke, Jacobs,
  Scholten, Schaefer-Prokop, Wille, Marchiano et~al.}]{ciompi2017towards}
\bibinfo{author}{Ciompi, F.}, \bibinfo{author}{Chung, K.},
  \bibinfo{author}{Van~Riel, S.J.}, \bibinfo{author}{Setio, A.A.A.},
  \bibinfo{author}{Gerke, P.K.}, \bibinfo{author}{Jacobs, C.},
  \bibinfo{author}{Scholten, E.T.}, \bibinfo{author}{Schaefer-Prokop, C.},
  \bibinfo{author}{Wille, M.M.}, \bibinfo{author}{Marchiano, A.}, et~al.,
  \bibinfo{year}{2017}.
\newblock \bibinfo{title}{Towards automatic pulmonary nodule management in lung
  cancer screening with deep learning}.
\newblock \bibinfo{journal}{Scientific {R}eports} \bibinfo{volume}{7},
  \bibinfo{pages}{46479}.
\bibitem[{Drozdzal et~al.(2016)Drozdzal, Vorontsov, Chartrand, Kadoury and
  Pal}]{drozdzal2016importance}
\bibinfo{author}{Drozdzal, M.}, \bibinfo{author}{Vorontsov, E.},
  \bibinfo{author}{Chartrand, G.}, \bibinfo{author}{Kadoury, S.},
  \bibinfo{author}{Pal, C.}, \bibinfo{year}{2016}.
\newblock \bibinfo{title}{The importance of skip connections in biomedical
  image segmentation}, in: \bibinfo{booktitle}{Deep learning and data labeling
  for medical applications}. \bibinfo{publisher}{Springer}, pp.
  \bibinfo{pages}{179--187}.
\bibitem[{Eaton-Rosen et~al.(2018)Eaton-Rosen, Bragman, Bisdas, Ourselin and
  Cardoso}]{eatonrosen2018safe}
\bibinfo{author}{Eaton-Rosen, Z.}, \bibinfo{author}{Bragman, F.},
  \bibinfo{author}{Bisdas, S.}, \bibinfo{author}{Ourselin, S.},
  \bibinfo{author}{Cardoso, M.J.}, \bibinfo{year}{2018}.
\newblock \bibinfo{title}{Towards safe deep learning: accurately quantifying
  biomarker uncertainty in neural network predictions}.
\newblock \bibinfo{journal}{arXiv preprint arXiv:1806.08640}
  \href{http://arxiv.org/abs/1806.08640}{\tt arXiv:1806.08640}.
\bibitem[{Elazab et~al.(2020)Elazab, Wang, Gardezi, Bai, Hu, Wang, Chang and
  Lei}]{elazab2020gp}
\bibinfo{author}{Elazab, A.}, \bibinfo{author}{Wang, C.},
  \bibinfo{author}{Gardezi, S.J.S.}, \bibinfo{author}{Bai, H.},
  \bibinfo{author}{Hu, Q.}, \bibinfo{author}{Wang, T.}, \bibinfo{author}{Chang,
  C.}, \bibinfo{author}{Lei, B.}, \bibinfo{year}{2020}.
\newblock \bibinfo{title}{{GP-GAN}: Brain tumor growth prediction using stacked
  3{D} generative adversarial networks from longitudinal {MR} {I}mages}.
\newblock \bibinfo{journal}{Neural {N}etworks} .
\bibitem[{Gal and Ghahramani(2016)}]{gal2016bayesian}
\bibinfo{author}{Gal, Y.}, \bibinfo{author}{Ghahramani, Z.},
  \bibinfo{year}{2016}.
\newblock \bibinfo{title}{Bayesian convolutional neural networks with
  {B}ernoulli approximate variational inference}.
\newblock \bibinfo{journal}{arXiv preprint arXiv:1506.02158}
  \href{http://arxiv.org/abs/1506.02158}{\tt arXiv:1506.02158}.
\bibitem[{Goodfellow(2016)}]{goodfellow2016nips}
\bibinfo{author}{Goodfellow, I.}, \bibinfo{year}{2016}.
\newblock \bibinfo{title}{Nips 2016 tutorial: Generative adversarial networks}.
\newblock \bibinfo{journal}{arXiv preprint arXiv:1701.00160} .
\bibitem[{Goodfellow et~al.(2014)Goodfellow, Pouget-Abadie, Mirza, Xu,
  Warde-Farley, Ozair, Courville and Bengio}]{goodfellow2014generative}
\bibinfo{author}{Goodfellow, I.}, \bibinfo{author}{Pouget-Abadie, J.},
  \bibinfo{author}{Mirza, M.}, \bibinfo{author}{Xu, B.},
  \bibinfo{author}{Warde-Farley, D.}, \bibinfo{author}{Ozair, S.},
  \bibinfo{author}{Courville, A.}, \bibinfo{author}{Bengio, Y.},
  \bibinfo{year}{2014}.
\newblock \bibinfo{title}{Generative adversarial nets}, in:
  \bibinfo{booktitle}{Advances in {N}eural {I}nformation {P}rocessing
  {S}ystems}, pp. \bibinfo{pages}{2672--2680}.
\bibitem[{Han et~al.(2018)Han, Heuvelmans, Vliegenthart, Rook, Dorrius,
  De~Jonge, Walter, van Ooijen, De~Koning and Oudkerk}]{han2018influence}
\bibinfo{author}{Han, D.}, \bibinfo{author}{Heuvelmans, M.A.},
  \bibinfo{author}{Vliegenthart, R.}, \bibinfo{author}{Rook, M.},
  \bibinfo{author}{Dorrius, M.D.}, \bibinfo{author}{De~Jonge, G.J.},
  \bibinfo{author}{Walter, J.E.}, \bibinfo{author}{van Ooijen, P.M.},
  \bibinfo{author}{De~Koning, H.J.}, \bibinfo{author}{Oudkerk, M.},
  \bibinfo{year}{2018}.
\newblock \bibinfo{title}{Influence of lung nodule margin on volume-and
  diameter-based reader variability in {CT} lung cancer screening}.
\newblock \bibinfo{journal}{The {B}ritish {J}ournal of {R}adiology}
  \bibinfo{volume}{91}, \bibinfo{pages}{20170405}.
\bibitem[{He et~al.(2016)He, Zhang, Ren and Sun}]{he2016identity}
\bibinfo{author}{He, K.}, \bibinfo{author}{Zhang, X.}, \bibinfo{author}{Ren,
  S.}, \bibinfo{author}{Sun, J.}, \bibinfo{year}{2016}.
\newblock \bibinfo{title}{Identity mappings in deep residual networks}, in:
  \bibinfo{booktitle}{European {C}onference on {C}omputer {V}ision},
  \bibinfo{organization}{Springer}. pp. \bibinfo{pages}{630--645}.
\bibitem[{Hochreiter and Schmidhuber(1997)}]{hochreiter1997long}
\bibinfo{author}{Hochreiter, S.}, \bibinfo{author}{Schmidhuber, J.},
  \bibinfo{year}{1997}.
\newblock \bibinfo{title}{Long short-term memory}.
\newblock \bibinfo{journal}{Neural {C}omputation} \bibinfo{volume}{9},
  \bibinfo{pages}{1735--1780}.
\bibitem[{Hu et~al.(2019)Hu, Worrall, Knegt, Veeling, Huisman and
  Welling}]{hu2019supervised}
\bibinfo{author}{Hu, S.}, \bibinfo{author}{Worrall, D.},
  \bibinfo{author}{Knegt, S.}, \bibinfo{author}{Veeling, B.},
  \bibinfo{author}{Huisman, H.}, \bibinfo{author}{Welling, M.},
  \bibinfo{year}{2019}.
\newblock \bibinfo{title}{Supervised uncertainty quantification for
  segmentation with multiple annotations}, in:
  \bibinfo{booktitle}{International {C}onference on {M}edical {I}mage
  {C}omputing and {C}omputer-{A}ssisted {I}ntervention},
  \bibinfo{organization}{Springer}. pp. \bibinfo{pages}{137--145}.
\bibitem[{Huang et~al.(2019)Huang, Lin, Li, Tammemagi, Brock, Atkar-Khattra,
  Xu, Hu, Mayo, Schmidt et~al.}]{huang2019prediction}
\bibinfo{author}{Huang, P.}, \bibinfo{author}{Lin, C.T.}, \bibinfo{author}{Li,
  Y.}, \bibinfo{author}{Tammemagi, M.C.}, \bibinfo{author}{Brock, M.V.},
  \bibinfo{author}{Atkar-Khattra, S.}, \bibinfo{author}{Xu, Y.},
  \bibinfo{author}{Hu, P.}, \bibinfo{author}{Mayo, J.R.},
  \bibinfo{author}{Schmidt, H.}, et~al., \bibinfo{year}{2019}.
\newblock \bibinfo{title}{Prediction of lung cancer risk at follow-up screening
  with low-dose {CT}: a training and validation study of a deep learning
  method}.
\newblock \bibinfo{journal}{The {L}ancet {D}igital {H}ealth}
  \bibinfo{volume}{1}, \bibinfo{pages}{e353--e362}.
\bibitem[{Isola et~al.(2017)Isola, Zhu, Zhou and Efros}]{isola2017image}
\bibinfo{author}{Isola, P.}, \bibinfo{author}{Zhu, J.Y.},
  \bibinfo{author}{Zhou, T.}, \bibinfo{author}{Efros, A.A.},
  \bibinfo{year}{2017}.
\newblock \bibinfo{title}{Image-to-image translation with conditional
  adversarial networks}, in: \bibinfo{booktitle}{Proceedings of the IEEE
  {C}onference on {C}omputer {V}ision and {P}attern {R}ecognition}, pp.
  \bibinfo{pages}{1125--1134}.
\bibitem[{Katzmann et~al.(2018)Katzmann, Muehlberg, S{\"u}hling, Noerenberg,
  Holch, Heinemann and Gro{\ss}}]{katzmann2018predicting}
\bibinfo{author}{Katzmann, A.}, \bibinfo{author}{Muehlberg, A.},
  \bibinfo{author}{S{\"u}hling, M.}, \bibinfo{author}{Noerenberg, D.},
  \bibinfo{author}{Holch, J.W.}, \bibinfo{author}{Heinemann, V.},
  \bibinfo{author}{Gro{\ss}, H.M.}, \bibinfo{year}{2018}.
\newblock \bibinfo{title}{Predicting lesion growth and patient survival in
  colorectal cancer patients using deep neural networks}, in:
  \bibinfo{booktitle}{Medical {I}maging with {D}eep {L}earning}.
\newblock \URLprefix \url{https://openreview.net/forum?id=B1e_uuoif}.
\bibitem[{Kendall et~al.(2015)Kendall, Badrinarayanan and
  Cipolla}]{kendall2015bayesian}
\bibinfo{author}{Kendall, A.}, \bibinfo{author}{Badrinarayanan, V.},
  \bibinfo{author}{Cipolla, R.}, \bibinfo{year}{2015}.
\newblock \bibinfo{title}{Bayesian segnet: Model uncertainty in deep
  convolutional encoder-decoder architectures for scene understanding}.
\newblock \bibinfo{journal}{arXiv preprint arXiv:1511.02680} .
\bibitem[{Kingma and Ba(2014)}]{kingma2014adam}
\bibinfo{author}{Kingma, D.P.}, \bibinfo{author}{Ba, J.}, \bibinfo{year}{2014}.
\newblock \bibinfo{title}{Adam: A method for stochastic optimization}.
\newblock \bibinfo{journal}{arXiv preprint arXiv:1412.6980} .
\bibitem[{Kingma and Welling(2013)}]{kingma2013auto}
\bibinfo{author}{Kingma, D.P.}, \bibinfo{author}{Welling, M.},
  \bibinfo{year}{2013}.
\newblock \bibinfo{title}{Auto-encoding variational {B}ayes}.
\newblock \bibinfo{journal}{arXiv preprint arXiv:1312.6114} .
\bibitem[{van Klaveren et~al.(2009)van Klaveren, Oudkerk, Prokop, Scholten,
  Nackaerts, Vernhout, van Iersel, van~den Bergh, van't Westeinde, van~der
  Aalst et~al.}]{van2009management}
\bibinfo{author}{van Klaveren, R.J.}, \bibinfo{author}{Oudkerk, M.},
  \bibinfo{author}{Prokop, M.}, \bibinfo{author}{Scholten, E.T.},
  \bibinfo{author}{Nackaerts, K.}, \bibinfo{author}{Vernhout, R.},
  \bibinfo{author}{van Iersel, C.A.}, \bibinfo{author}{van~den Bergh, K.A.},
  \bibinfo{author}{van't Westeinde, S.}, \bibinfo{author}{van~der Aalst, C.},
  et~al., \bibinfo{year}{2009}.
\newblock \bibinfo{title}{Management of lung nodules detected by volume {CT}
  scanning}.
\newblock \bibinfo{journal}{New {E}ngland {J}ournal of {M}edicine}
  \bibinfo{volume}{361}, \bibinfo{pages}{2221--2229}.
\bibitem[{Kohl et~al.(2018)Kohl, Romera-Paredes, Meyer, De~Fauw, Ledsam,
  Maier-Hein, Eslami, Rezende and Ronneberger}]{kohl2018probabilistic}
\bibinfo{author}{Kohl, S.}, \bibinfo{author}{Romera-Paredes, B.},
  \bibinfo{author}{Meyer, C.}, \bibinfo{author}{De~Fauw, J.},
  \bibinfo{author}{Ledsam, J.R.}, \bibinfo{author}{Maier-Hein, K.},
  \bibinfo{author}{Eslami, S.A.}, \bibinfo{author}{Rezende, D.J.},
  \bibinfo{author}{Ronneberger, O.}, \bibinfo{year}{2018}.
\newblock \bibinfo{title}{A probabilistic {U}-{N}et for segmentation of
  ambiguous images}, in: \bibinfo{booktitle}{Advances in {N}eural {I}nformation
  {P}rocessing {S}ystems}, pp. \bibinfo{pages}{6965--6975}.
\bibitem[{Kohl et~al.(2019)Kohl, Romera-Paredes, Maier-Hein, Rezende, Eslami,
  Kohli, Zisserman and Ronneberger}]{kohl2019hierarchical}
\bibinfo{author}{Kohl, S.A.}, \bibinfo{author}{Romera-Paredes, B.},
  \bibinfo{author}{Maier-Hein, K.H.}, \bibinfo{author}{Rezende, D.J.},
  \bibinfo{author}{Eslami, S.}, \bibinfo{author}{Kohli, P.},
  \bibinfo{author}{Zisserman, A.}, \bibinfo{author}{Ronneberger, O.},
  \bibinfo{year}{2019}.
\newblock \bibinfo{title}{A hierarchical probabilistic {U}-{N}et for modeling
  multi-scale ambiguities}.
\newblock \bibinfo{journal}{arXiv preprint arXiv:1905.13077} .
\bibitem[{Leibig et~al.(2017)Leibig, Allken, Ayhan, Berens and
  Wahl}]{leibig2017leveraging}
\bibinfo{author}{Leibig, C.}, \bibinfo{author}{Allken, V.},
  \bibinfo{author}{Ayhan, M.S.}, \bibinfo{author}{Berens, P.},
  \bibinfo{author}{Wahl, S.}, \bibinfo{year}{2017}.
\newblock \bibinfo{title}{Leveraging uncertainty information from deep neural
  networks for disease detection}.
\newblock \bibinfo{journal}{Scientific {R}eports} \bibinfo{volume}{7},
  \bibinfo{pages}{1--14}.
\bibitem[{Li et~al.(2020a)Li, Tang, Chan, Zhou and Qian}]{li2020dc}
\bibinfo{author}{Li, M.}, \bibinfo{author}{Tang, H.}, \bibinfo{author}{Chan,
  M.D.}, \bibinfo{author}{Zhou, X.}, \bibinfo{author}{Qian, X.},
  \bibinfo{year}{2020}a.
\newblock \bibinfo{title}{{DC-AL-GAN}: pseudoprogression and true tumor
  progression of glioblastoma multiform image classification based on {DCGAN}
  and {A}lex{N}et}.
\newblock \bibinfo{journal}{Medical {P}hysics} \bibinfo{volume}{47},
  \bibinfo{pages}{1139--1150}.
\bibitem[{Li et~al.(2020b)Li, Yang, Xu, Xu, Ye, Tao, Xie and
  Liu}]{li2020learning}
\bibinfo{author}{Li, Y.}, \bibinfo{author}{Yang, J.}, \bibinfo{author}{Xu, Y.},
  \bibinfo{author}{Xu, J.}, \bibinfo{author}{Ye, X.}, \bibinfo{author}{Tao,
  G.}, \bibinfo{author}{Xie, X.}, \bibinfo{author}{Liu, G.},
  \bibinfo{year}{2020}b.
\newblock \bibinfo{title}{Learning tumor growth via follow-up volume prediction
  for lung nodules}, in: \bibinfo{booktitle}{International {C}onference on
  {M}edical {I}mage {C}omputing and {C}omputer-{A}ssisted {I}ntervention},
  \bibinfo{organization}{Springer}. pp. \bibinfo{pages}{508--517}.
\bibitem[{Lipkov{\'a} et~al.(2019)Lipkov{\'a}, Angelikopoulos, Wu, Alberts,
  Wiestler, Diehl, Preibisch, Pyka, Combs, Hadjidoukas
  et~al.}]{lipkova2019personalized}
\bibinfo{author}{Lipkov{\'a}, J.}, \bibinfo{author}{Angelikopoulos, P.},
  \bibinfo{author}{Wu, S.}, \bibinfo{author}{Alberts, E.},
  \bibinfo{author}{Wiestler, B.}, \bibinfo{author}{Diehl, C.},
  \bibinfo{author}{Preibisch, C.}, \bibinfo{author}{Pyka, T.},
  \bibinfo{author}{Combs, S.E.}, \bibinfo{author}{Hadjidoukas, P.}, et~al.,
  \bibinfo{year}{2019}.
\newblock \bibinfo{title}{Personalized radiotherapy design for glioblastoma:
  integrating mathematical tumor models, multimodal scans, and {B}ayesian
  inference}.
\newblock \bibinfo{journal}{IEEE {T}ransactions on {M}edical {I}maging}
  \bibinfo{volume}{38}, \bibinfo{pages}{1875--1884}.
\bibitem[{Luong et~al.(2015)Luong, Pham and Manning}]{luong2015effective}
\bibinfo{author}{Luong, M.T.}, \bibinfo{author}{Pham, H.},
  \bibinfo{author}{Manning, C.D.}, \bibinfo{year}{2015}.
\newblock \bibinfo{title}{Effective approaches to attention-based neural
  machine translation}.
\newblock \bibinfo{journal}{arXiv preprint arXiv:1508.04025} .
\bibitem[{MacMahon et~al.(2017)MacMahon, Naidich, Goo, Lee, Leung, Mayo, Mehta,
  Ohno, Powell, Prokop et~al.}]{macmahon2017guidelines}
\bibinfo{author}{MacMahon, H.}, \bibinfo{author}{Naidich, D.P.},
  \bibinfo{author}{Goo, J.M.}, \bibinfo{author}{Lee, K.S.},
  \bibinfo{author}{Leung, A.N.}, \bibinfo{author}{Mayo, J.R.},
  \bibinfo{author}{Mehta, A.C.}, \bibinfo{author}{Ohno, Y.},
  \bibinfo{author}{Powell, C.A.}, \bibinfo{author}{Prokop, M.}, et~al.,
  \bibinfo{year}{2017}.
\newblock \bibinfo{title}{Guidelines for management of incidental pulmonary
  nodules detected on {CT} images: from the {F}leischner society 2017}.
\newblock \bibinfo{journal}{Radiology} \bibinfo{volume}{284},
  \bibinfo{pages}{228--243}.
\bibitem[{Messay et~al.(2015)Messay, Hardie and
  Tuinstra}]{messay2015segmentation}
\bibinfo{author}{Messay, T.}, \bibinfo{author}{Hardie, R.C.},
  \bibinfo{author}{Tuinstra, T.R.}, \bibinfo{year}{2015}.
\newblock \bibinfo{title}{Segmentation of pulmonary nodules in computed
  tomography using a regression neural network approach and its application to
  the lung image database consortium and image database resource initiative
  dataset}.
\newblock \bibinfo{journal}{Medical {I}mage {A}nalysis} \bibinfo{volume}{22},
  \bibinfo{pages}{48--62}.
\bibitem[{Milletari et~al.(2016)Milletari, Navab and Ahmadi}]{milletari2016v}
\bibinfo{author}{Milletari, F.}, \bibinfo{author}{Navab, N.},
  \bibinfo{author}{Ahmadi, S.A.}, \bibinfo{year}{2016}.
\newblock \bibinfo{title}{V-net: Fully convolutional neural networks for
  volumetric medical image segmentation}, in: \bibinfo{booktitle}{2016 fourth
  international conference on 3{D} vision (3DV)}, \bibinfo{organization}{IEEE}.
  pp. \bibinfo{pages}{565--571}.
\bibitem[{Nair et~al.(2020)Nair, Precup, Arnold and Arbel}]{nair2020exploring}
\bibinfo{author}{Nair, T.}, \bibinfo{author}{Precup, D.},
  \bibinfo{author}{Arnold, D.L.}, \bibinfo{author}{Arbel, T.},
  \bibinfo{year}{2020}.
\newblock \bibinfo{title}{Exploring uncertainty measures in deep networks for
  multiple sclerosis lesion detection and segmentation}.
\newblock \bibinfo{journal}{Medical {I}mage {A}nalysis} \bibinfo{volume}{59}.
\bibitem[{Najafabadi et~al.(2015)Najafabadi, Villanustre, Khoshgoftaar, Seliya,
  Wald and Muharemagic}]{najafabadi2015deep}
\bibinfo{author}{Najafabadi, M.M.}, \bibinfo{author}{Villanustre, F.},
  \bibinfo{author}{Khoshgoftaar, T.M.}, \bibinfo{author}{Seliya, N.},
  \bibinfo{author}{Wald, R.}, \bibinfo{author}{Muharemagic, E.},
  \bibinfo{year}{2015}.
\newblock \bibinfo{title}{Deep learning applications and challenges in big data
  analytics}.
\newblock \bibinfo{journal}{Journal of {B}ig {D}ata} \bibinfo{volume}{2},
  \bibinfo{pages}{1}.
\bibitem[{Oksuz et~al.(2020)Oksuz, Cam, Kalkan and Akbas}]{oksuz2020imbalance}
\bibinfo{author}{Oksuz, K.}, \bibinfo{author}{Cam, B.C.},
  \bibinfo{author}{Kalkan, S.}, \bibinfo{author}{Akbas, E.},
  \bibinfo{year}{2020}.
\newblock \bibinfo{title}{Imbalance problems in object detection: A review}.
\newblock \bibinfo{journal}{IEEE {T}ransactions on {P}attern {A}nalysis and
  {M}achine {I}ntelligence} .
\bibitem[{Oktay et~al.(2018)Oktay, Schlemper, Folgoc, Lee, Heinrich, Misawa,
  Mori, McDonagh, Hammerla, Kainz, Glocker and Rueckert}]{oktay2018attention}
\bibinfo{author}{Oktay, O.}, \bibinfo{author}{Schlemper, J.},
  \bibinfo{author}{Folgoc, L.L.}, \bibinfo{author}{Lee, M.},
  \bibinfo{author}{Heinrich, M.}, \bibinfo{author}{Misawa, K.},
  \bibinfo{author}{Mori, K.}, \bibinfo{author}{McDonagh, S.},
  \bibinfo{author}{Hammerla, N.Y.}, \bibinfo{author}{Kainz, B.},
  \bibinfo{author}{Glocker, B.}, \bibinfo{author}{Rueckert, D.},
  \bibinfo{year}{2018}.
\newblock \bibinfo{title}{Attention {U}-{N}et: Learning where to look for the
  pancreas}.
\newblock \href{http://arxiv.org/abs/1804.03999}{\tt arXiv:1804.03999}.
\bibitem[{Osband(2016)}]{Osband2016Risk}
\bibinfo{author}{Osband, I.}, \bibinfo{year}{2016}.
\newblock \bibinfo{title}{Risk versus uncertainty in deep learning: Bayes,
  bootstrap and the dangers of dropout}, in: \bibinfo{booktitle}{{NIPS}
  {W}orkshop on {B}ayesian {D}eep {L}earning}.
\bibitem[{Petersen et~al.(2019)Petersen, J{\"a}ger, Isensee, Kohl, Neuberger,
  Wick, Debus, Heiland, Bendszus, Kickingereder et~al.}]{petersen2019deep}
\bibinfo{author}{Petersen, J.}, \bibinfo{author}{J{\"a}ger, P.F.},
  \bibinfo{author}{Isensee, F.}, \bibinfo{author}{Kohl, S.A.},
  \bibinfo{author}{Neuberger, U.}, \bibinfo{author}{Wick, W.},
  \bibinfo{author}{Debus, J.}, \bibinfo{author}{Heiland, S.},
  \bibinfo{author}{Bendszus, M.}, \bibinfo{author}{Kickingereder, P.}, et~al.,
  \bibinfo{year}{2019}.
\newblock \bibinfo{title}{Deep probabilistic modeling of glioma growth}, in:
  \bibinfo{booktitle}{International {C}onference on {M}edical {I}mage
  {C}omputing and {C}omputer-{A}ssisted {I}ntervention},
  \bibinfo{organization}{Springer}. pp. \bibinfo{pages}{806--814}.
\bibitem[{Popovic and Thomas(2017)}]{popovic2017assessing}
\bibinfo{author}{Popovic, Z.B.}, \bibinfo{author}{Thomas, J.D.},
  \bibinfo{year}{2017}.
\newblock \bibinfo{title}{Assessing observer variability: a user’s guide}.
\newblock \bibinfo{journal}{Cardiovascular {D}iagnosis and {T}herapy}
  \bibinfo{volume}{7}, \bibinfo{pages}{317}.
\bibitem[{Prasoon et~al.(2013)Prasoon, Petersen, Igel, Lauze, Dam and
  Nielsen}]{prasoon2013deep}
\bibinfo{author}{Prasoon, A.}, \bibinfo{author}{Petersen, K.},
  \bibinfo{author}{Igel, C.}, \bibinfo{author}{Lauze, F.},
  \bibinfo{author}{Dam, E.}, \bibinfo{author}{Nielsen, M.},
  \bibinfo{year}{2013}.
\newblock \bibinfo{title}{Deep feature learning for knee cartilage segmentation
  using a triplanar convolutional neural network}, in:
  \bibinfo{booktitle}{International {C}onference on {M}edical {I}mage
  {C}omputing and {C}omputer-{A}ssisted {I}ntervention},
  \bibinfo{organization}{Springer}. pp. \bibinfo{pages}{246--253}.
\bibitem[{Rachmadi et~al.(2020)Rachmadi, Vald{\'e}s-Hern{\'a}ndez, Makin,
  Wardlaw and Komura}]{rachmadi2020automatic}
\bibinfo{author}{Rachmadi, M.F.}, \bibinfo{author}{Vald{\'e}s-Hern{\'a}ndez,
  M.d.C.}, \bibinfo{author}{Makin, S.}, \bibinfo{author}{Wardlaw, J.},
  \bibinfo{author}{Komura, T.}, \bibinfo{year}{2020}.
\newblock \bibinfo{title}{Automatic spatial estimation of white matter
  hyperintensities evolution in brain {MRI} using disease evolution predictor
  deep neural networks}.
\newblock \bibinfo{journal}{Medical {I}mage {A}nalysis} .
\bibitem[{Rafael-Palou et~al.(2020)Rafael-Palou, Aubanell, Bonavita, Ceresa,
  Piella, Ribas and Ballester}]{rafael2020re}
\bibinfo{author}{Rafael-Palou, X.}, \bibinfo{author}{Aubanell, A.},
  \bibinfo{author}{Bonavita, I.}, \bibinfo{author}{Ceresa, M.},
  \bibinfo{author}{Piella, G.}, \bibinfo{author}{Ribas, V.},
  \bibinfo{author}{Ballester, M.A.G.}, \bibinfo{year}{2020}.
\newblock \bibinfo{title}{Re-identification and growth detection of pulmonary
  nodules without image registration using 3{D} siamese neural networks}.
\newblock \bibinfo{journal}{Medical {I}mage {A}nalysis} .
\bibitem[{Ravi et~al.(2019)Ravi, Blumberg, Mengoudi, Xu, Alexander and
  Oxtoby}]{ravi2019degenerative}
\bibinfo{author}{Ravi, D.}, \bibinfo{author}{Blumberg, S.B.},
  \bibinfo{author}{Mengoudi, K.}, \bibinfo{author}{Xu, M.},
  \bibinfo{author}{Alexander, D.C.}, \bibinfo{author}{Oxtoby, N.P.},
  \bibinfo{year}{2019}.
\newblock \bibinfo{title}{Degenerative adversarial neuroimage nets for {4D}
  simulations: Application in longitudinal {MRI}}.
\newblock \bibinfo{journal}{arXiv preprint arXiv:1912.01526} .
\bibitem[{Rezende et~al.(2014)Rezende, Mohamed and
  Wierstra}]{rezende2014stochastic}
\bibinfo{author}{Rezende, D.J.}, \bibinfo{author}{Mohamed, S.},
  \bibinfo{author}{Wierstra, D.}, \bibinfo{year}{2014}.
\newblock \bibinfo{title}{Stochastic backpropagation and approximate inference
  in deep generative models}.
\newblock \bibinfo{journal}{arXiv preprint arXiv:1401.4082} .
\bibitem[{Ronneberger et~al.(2015)Ronneberger, Fischer and
  Brox}]{ronneberger2015u}
\bibinfo{author}{Ronneberger, O.}, \bibinfo{author}{Fischer, P.},
  \bibinfo{author}{Brox, T.}, \bibinfo{year}{2015}.
\newblock \bibinfo{title}{{U}-{N}et: Convolutional networks for biomedical
  image segmentation}, in: \bibinfo{booktitle}{International {C}onference on
  {M}edical {I}mage {C}omputing and {C}omputer-{A}ssisted {I}ntervention},
  \bibinfo{organization}{Springer}. pp. \bibinfo{pages}{234--241}.
\bibitem[{Roy et~al.(2018)Roy, Conjeti, Navab and Wachinger}]{roy2018inherent}
\bibinfo{author}{Roy, A.G.}, \bibinfo{author}{Conjeti, S.},
  \bibinfo{author}{Navab, N.}, \bibinfo{author}{Wachinger, C.},
  \bibinfo{year}{2018}.
\newblock \bibinfo{title}{Inherent brain segmentation quality control from
  fully convnet {M}onte {C}arlo sampling}, in:
  \bibinfo{booktitle}{International {C}onference on {M}edical {I}mage
  {C}omputing and {C}omputer-{A}ssisted {I}ntervention},
  \bibinfo{organization}{Springer}. pp. \bibinfo{pages}{664--672}.
\bibitem[{Sarapata and de~Pillis(2014)}]{sarapata2014comparison}
\bibinfo{author}{Sarapata, E.A.}, \bibinfo{author}{de~Pillis, L.},
  \bibinfo{year}{2014}.
\newblock \bibinfo{title}{A comparison and catalog of intrinsic tumor growth
  models}.
\newblock \bibinfo{journal}{Bulletin of {M}athematical {B}iology}
  \bibinfo{volume}{76}, \bibinfo{pages}{2010--2024}.
\bibitem[{Setio et~al.(2017)Setio, Traverso, De~Bel, Berens, van~den Bogaard,
  Cerello, Chen, Dou, Fantacci, Geurts et~al.}]{setio2017validation}
\bibinfo{author}{Setio, A.A.A.}, \bibinfo{author}{Traverso, A.},
  \bibinfo{author}{De~Bel, T.}, \bibinfo{author}{Berens, M.S.},
  \bibinfo{author}{van~den Bogaard, C.}, \bibinfo{author}{Cerello, P.},
  \bibinfo{author}{Chen, H.}, \bibinfo{author}{Dou, Q.},
  \bibinfo{author}{Fantacci, M.E.}, \bibinfo{author}{Geurts, B.}, et~al.,
  \bibinfo{year}{2017}.
\newblock \bibinfo{title}{Validation, comparison, and combination of algorithms
  for automatic detection of pulmonary nodules in computed tomography images:
  the {LUNA}16 challenge}.
\newblock \bibinfo{journal}{Medical {I}mage {A}nalysis} \bibinfo{volume}{42},
  \bibinfo{pages}{1--13}.
\bibitem[{Shi et~al.(2015)Shi, Chen, Wang, Yeung, Wong and
  Woo}]{shi2015convolutional}
\bibinfo{author}{Shi, X.}, \bibinfo{author}{Chen, Z.}, \bibinfo{author}{Wang,
  H.}, \bibinfo{author}{Yeung, D.Y.}, \bibinfo{author}{Wong, W.K.},
  \bibinfo{author}{Woo, W.c.}, \bibinfo{year}{2015}.
\newblock \bibinfo{title}{Convolutional {LSTM} network: A machine learning
  approach for precipitation nowcasting}.
\newblock \bibinfo{journal}{Advances in {N}eural {I}nformation {P}rocessing
  {S}ystems} \bibinfo{volume}{28}, \bibinfo{pages}{802--810}.
\bibitem[{Shridhar et~al.(2019)Shridhar, Laumann and
  Liwicki}]{shridhar2019uncertainty}
\bibinfo{author}{Shridhar, K.}, \bibinfo{author}{Laumann, F.},
  \bibinfo{author}{Liwicki, M.}, \bibinfo{year}{2019}.
\newblock \bibinfo{title}{Uncertainty estimations by softplus normalization in
  {B}ayesian convolutional neural networks with variational inference}.
\newblock \bibinfo{journal}{arXiv preprint arXiv:1806.05978}
  \href{http://arxiv.org/abs/1806.05978}{\tt arXiv:1806.05978}.
\bibitem[{Sohn et~al.(2015)Sohn, Lee and Yan}]{sohn2015learning}
\bibinfo{author}{Sohn, K.}, \bibinfo{author}{Lee, H.}, \bibinfo{author}{Yan,
  X.}, \bibinfo{year}{2015}.
\newblock \bibinfo{title}{Learning structured output representation using deep
  conditional generative models}, in: \bibinfo{booktitle}{Advances in {N}eural
  {I}nformation {P}rocessing {S}ystems}, pp. \bibinfo{pages}{3483--3491}.
\bibitem[{Swanson et~al.(2002)Swanson, Alvord and
  Murray}]{swanson2002quantifying}
\bibinfo{author}{Swanson, K.R.}, \bibinfo{author}{Alvord, E.C.},
  \bibinfo{author}{Murray, J.}, \bibinfo{year}{2002}.
\newblock \bibinfo{title}{Quantifying efficacy of chemotherapy of brain tumors
  with homogeneous and heterogeneous drug delivery}.
\newblock \bibinfo{journal}{Acta {B}iotheoretica} \bibinfo{volume}{50},
  \bibinfo{pages}{223--237}.
\bibitem[{Sz{\'e}kely and Rizzo(2013)}]{szekely2013energy}
\bibinfo{author}{Sz{\'e}kely, G.J.}, \bibinfo{author}{Rizzo, M.L.},
  \bibinfo{year}{2013}.
\newblock \bibinfo{title}{Energy statistics: A class of statistics based on
  distances}.
\newblock \bibinfo{journal}{Journal of {S}tatistical {P}lanning and
  {I}nference} \bibinfo{volume}{143}, \bibinfo{pages}{1249--1272}.
\bibitem[{Talkington and Durrett(2015)}]{talkington2015estimating}
\bibinfo{author}{Talkington, A.}, \bibinfo{author}{Durrett, R.},
  \bibinfo{year}{2015}.
\newblock \bibinfo{title}{Estimating tumor growth rates in vivo}.
\newblock \bibinfo{journal}{Bulletin of {M}athematical {B}iology}
  \bibinfo{volume}{77}, \bibinfo{pages}{1934--1954}.
\bibitem[{Wang et~al.(2019)Wang, Rimner, Hu, Tyagi, Jiang, Yorke, Riyahi,
  Mageras, Deasy and Zhang}]{wang2019toward}
\bibinfo{author}{Wang, C.}, \bibinfo{author}{Rimner, A.}, \bibinfo{author}{Hu,
  Y.C.}, \bibinfo{author}{Tyagi, N.}, \bibinfo{author}{Jiang, J.},
  \bibinfo{author}{Yorke, E.}, \bibinfo{author}{Riyahi, S.},
  \bibinfo{author}{Mageras, G.}, \bibinfo{author}{Deasy, J.O.},
  \bibinfo{author}{Zhang, P.}, \bibinfo{year}{2019}.
\newblock \bibinfo{title}{Toward predicting the evolution of lung tumors during
  radiotherapy observed on a longitudinal mr imaging study via a deep learning
  algorithm}.
\newblock \bibinfo{journal}{Medical {P}hysics} \bibinfo{volume}{46},
  \bibinfo{pages}{4699--4707}.
\bibitem[{Wong et~al.(2015)Wong, Summers, Kebebew and Yao}]{wong2015tumor}
\bibinfo{author}{Wong, K.C.}, \bibinfo{author}{Summers, R.M.},
  \bibinfo{author}{Kebebew, E.}, \bibinfo{author}{Yao, J.},
  \bibinfo{year}{2015}.
\newblock \bibinfo{title}{Tumor growth prediction with reaction-diffusion and
  hyperelastic biomechanical model by physiological data fusion}.
\newblock \bibinfo{journal}{Medical {I}mage {A}nalysis} \bibinfo{volume}{25},
  \bibinfo{pages}{72--85}.
\bibitem[{Wong et~al.(2016)Wong, Summers, Kebebew and Yao}]{wong2016pancreatic}
\bibinfo{author}{Wong, K.C.}, \bibinfo{author}{Summers, R.M.},
  \bibinfo{author}{Kebebew, E.}, \bibinfo{author}{Yao, J.},
  \bibinfo{year}{2016}.
\newblock \bibinfo{title}{Pancreatic tumor growth prediction with
  elastic-growth decomposition, image-derived motion, and {FDM-FEM} coupling}.
\newblock \bibinfo{journal}{IEEE {T}ransactions on {M}edical {I}maging}
  \bibinfo{volume}{36}, \bibinfo{pages}{111--123}.
\bibitem[{Zhang et~al.(2017)Zhang, Lu, Summers, Kebebew and
  Yao}]{zhang2017convolutional}
\bibinfo{author}{Zhang, L.}, \bibinfo{author}{Lu, L.},
  \bibinfo{author}{Summers, R.M.}, \bibinfo{author}{Kebebew, E.},
  \bibinfo{author}{Yao, J.}, \bibinfo{year}{2017}.
\newblock \bibinfo{title}{Convolutional invasion and expansion networks for
  tumor growth prediction}.
\newblock \bibinfo{journal}{IEEE {T}ransactions on {M}edical {I}maging}
  \bibinfo{volume}{37}, \bibinfo{pages}{638--648}.
\bibitem[{Zhang et~al.(2019)Zhang, Lu, Wang, Zhu, Bagheri, Summers and
  Yao}]{zhang2019spatio}
\bibinfo{author}{Zhang, L.}, \bibinfo{author}{Lu, L.}, \bibinfo{author}{Wang,
  X.}, \bibinfo{author}{Zhu, R.M.}, \bibinfo{author}{Bagheri, M.},
  \bibinfo{author}{Summers, R.M.}, \bibinfo{author}{Yao, J.},
  \bibinfo{year}{2019}.
\newblock \bibinfo{title}{Spatio-temporal convolutional {LSTM}s for tumor
  growth prediction by learning {4D} longitudinal patient data}.
\newblock \bibinfo{journal}{IEEE {T}ransactions on {M}edical {I}maging}
  \bibinfo{volume}{39}, \bibinfo{pages}{1114--1126}.
\bibitem[{Zhang et~al.(2018)Zhang, Liu and Wang}]{zhang2018road}
\bibinfo{author}{Zhang, Z.}, \bibinfo{author}{Liu, Q.}, \bibinfo{author}{Wang,
  Y.}, \bibinfo{year}{2018}.
\newblock \bibinfo{title}{Road extraction by deep residual {U}-{N}et}.
\newblock \bibinfo{journal}{IEEE {G}eoscience and {R}emote {S}ensing {L}etters}
  \bibinfo{volume}{15}, \bibinfo{pages}{749--753}.

\end{thebibliography}
\end{document}